\def\year{2021}\relax
\documentclass[twocolumn]{autart} 
\usepackage{times}  
\usepackage{helvet} 
\usepackage{courier}  
\usepackage[hyphens]{url}  
\usepackage{graphicx} 
\usepackage{natbib}  
\usepackage{caption} 
\newcommand{\bx}{{\bf x}}
\newcommand{\bz}{{\bf z}}
\newcommand{\bs}{{\bf s}}
\newcommand{\ba}{{\bf a}}
\newcommand{\bb}{{\bf b}}
\newcommand{\bm}{{\bf m}}
\newcommand{\bl}{{\boldsymbol \ell}}

\usepackage{thmtools,thm-restate}
\usepackage{subcaption}
\usepackage{standalone}
\usepackage{nicefrac}
\usepackage{arydshln}
\usepackage{wrapfig}
\usepackage{tikz}
\usetikzlibrary{intersections}
\usetikzlibrary{decorations.pathreplacing}
\usetikzlibrary{automata,arrows,positioning,calc}
\usepackage[fleqn]{amsmath}
\usepackage{amssymb}
\usepackage{amsfonts}
\DeclareMathOperator*{\argmin}{argmin}
\usetikzlibrary{shapes.geometric}
\usetikzlibrary{decorations.pathmorphing}
\tikzset{
   ragged border/.style={ decoration={random steps, segment length=1mm, amplitude=.75mm},
           decorate,
   }
}

\newtheorem{assumption}{\bf Assumption}

\newtheorem{remark}{\bf Remark}
\newtheorem{lemma}{\bf Lemma}
\allowdisplaybreaks

\begin{document}

\begin{frontmatter}

\title{Simultaneous Perception-Action Design \\ via Invariant Finite Belief Sets\thanksref{footnoteinfo}}

\thanks[footnoteinfo]{This paper was not presented at any IFAC 
meeting. Corresponding author M.~Hibbard.}

\author[UT]{Michael Hibbard}\ead{mwhibbard@utexas.edu}, { }
\author[UT]{Takashi Tanaka}\ead{ttanaka@utexas.edu}, { }
\author[UT]{Ufuk Topcu}\ead{utopcu@utexas.edu}

\address[UT]{Department of Aerospace Engineering and Engineering Mechanics University of Texas at Austin}

\begin{keyword}                 
Perception and sensing; Planning; Optimization under uncertainties.  
\end{keyword}  

\begin{abstract}
Although perception is an increasingly dominant portion of the overall computational cost for autonomous systems, only a fraction of the information perceived is likely to be relevant to the current task. To alleviate these perception costs, we develop a novel simultaneous perception-action design framework wherein an agent senses only the task-relevant information. This formulation differs from that of a partially observable Markov decision process, since the agent is free to synthesize not only its policy for action selection but also its belief-dependent observation function.
The method enables the agent to balance its perception costs with those incurred by operating in its environment. To obtain a computationally tractable solution, we approximate the value function using a novel \textit{method of invariant finite belief sets}, wherein the agent acts exclusively on a finite subset of the continuous belief space. We solve the approximate problem through value iteration in which a linear program is solved individually for each belief state in the set, in each iteration. Finally, we prove that the value functions, under an assumption on their structure, converge to their continuous state-space values as the sample density increases. 
\end{abstract}

\end{frontmatter}

\section{Introduction}\label{sec:Introduction}

Evolution has driven biological organisms to strike a balance between the conflicting desire of utilizing all available information in order to make a decision and the desire of minimizing the cost of perceiving that information from the environment.
Such perception costs are not negligible: studies of the human brain, for instance, have shown that it constitutes 20\% of our resting energy consumption rate, of which 50\% is associated with signaling (\cite{attwell2001energy}).
In order to optimally utilize only a portion of the available information, it is believed that organisms have evolved to strategically perceive only the \textit{task-relevant} information from their environment 
(\cite{berry1999anticipation,egner2005cognitive}).
%
The mammalian visual cortex, for example, is sensitive to only particular features over a small region of the visual field (\cite{hubel1968receptive}). 

Perception costs have likewise become a bottleneck in many engineering applications. For example, experimental results have shown that over 94\% of the computational time in autonomous driving is allocated to perception (\cite{lin2018architectural}). Although state-of-the-art accelerator platforms like GPUs are effective for latency reduction, their power consumption is significant enough to degrade a vehicle’s driving range. To alleviate these issues, \cite{censi2015power} argues that sensor hardware should extract only task-relevant information. Although intuitively appealing, such an idea is difficult to implement, as what constitutes ``task-relevant" information is difficult to define.

To provide a methodological foundation for task-relevant sensing, we propose a simultaneous perception-action design (SPADE) framework based on the standard Markov decision process (MDP) formulation (\cite{puterman2014markov}) with a novel information-theoretic perception cost. The perception cost penalizes information flowing from the sensor to the down-stream decision-making unit. This framework allows for the synthesis of a sensing mechanism that extracts the minimum amount of task-relevant information from the underlying state of the controlled Markov chain.

The use of information theory to model perception costs has previously been studied. Viewing an agent as a communication channel, \cite{sims2003implications} proposes a model penalizing the mutual information between the state of the environment and the agent’s action. Likewise, \cite{sims2016rate} proposed using rate-distortion theory to characterize perception costs.
The analogy between the perception-action cycle and a communication channel was also studied in \cite{tishby2011information}, where algorithms to synthesize the optimal trade-off between the cost-to-go and the information-to-go were proposed.
Alternatively, \cite{ortega2013thermodynamics} studied the problem of rational inattention through the lens of thermodynamics, where information processing costs are characterized through differences in free energy.
Rational inattention was also studied in \cite{shafieepoorfard2016rationally}, which provided theoretical results connecting controllers subject to information constraints to rate-distortion theory.
Following \cite{massey1990causality}, we use the information-theoretic concept of \textit{directed information} to model the agent's perception costs. To our knowledge, this paper is the first to apply directed information to the study of optimal perception.

The proposed SPADE framework is reminiscent of the existing research on \textit{active perception} (\cite{aloimonos2013active,bajcsy2018revisiting}). In active perception problems, an agent (or group of agents, as in \cite{spaan2008cooperative}) seeks to take actions that lead to desired observations. As opposed to existing works on active perception, the SPADE framework allows an agent to additionally synthesize its own perception mechanism, rather than acting in such a way as to exploit its existing perception mechanism.

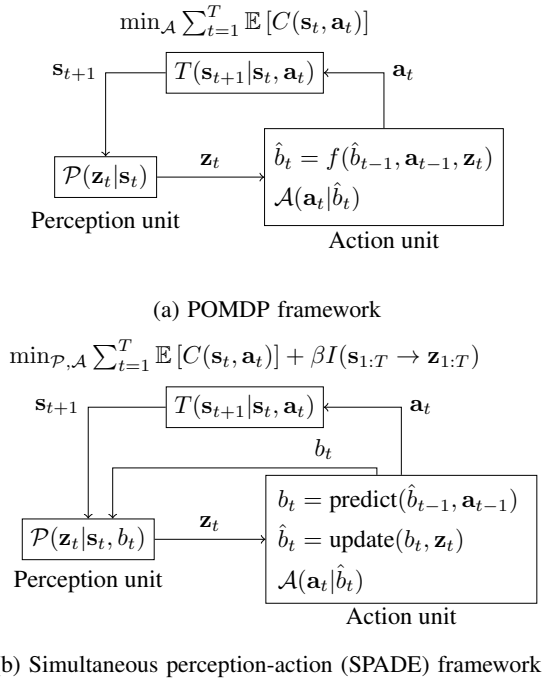
\begin{figure}
    \begin{subfigure}[b]{0.49\textwidth}
        \centering \scalebox{0.925}{
        \begin{tikzpicture}

\clip (-3.75,-2.9) rectangle + (8.25,4.15);  

\node (costFunc) at (0,0.75) {$\min_{\mathcal{A}}\sum_{t=1}^{T}\mathbb{E}\left[ C(\bs_{t},\ba_{t}) \right]$};

\node (T) at (0,0) [rectangle,draw] {$T(\bs_{t+1}|\bs_{t},\ba_{t})$};

\node (P) at (-2,-1.5) [rectangle,draw] {$\mathcal{P}(\bz_{t}|\bs_{t})$};
\node (P_label) at (-2,-2.15) {Perception unit};

\node[rectangle,draw] (A) at (2,-1.5)  {\hspace*{-0.7cm} \begin{tabular}{lll}
     & $\hat{b}_{t}=f(\hat{b}_{t-1},\ba_{t-1},\bz_{t})$ & \\
     & $\mathcal{A}(\ba_{t}|\hat{b}_{t})$ &
\end{tabular} \hspace*{-0.7cm}};

\node[align=left] (A_label) at (2,-2.4) {Action unit};

\draw[->] (P) -- node[above] {$\bz_{t}$} (A);
\draw[->] (A) |- node[right] {$\ba_{t}$} (T);
\draw[->] (T) -| node[left] {$\bs_{t+1}$} (P);

\end{tikzpicture}}
        \caption{POMDP framework}
        \label{fig:POMDP_diagram}
    \end{subfigure}
    \begin{subfigure}[b]{0.49\textwidth}
        \centering \scalebox{0.925}{
        \begin{tikzpicture}

\clip (-3.75,-3.2) rectangle + (8.25,4.35);
    

\node (costFunc) at (0,0.75) {$\min_{\mathcal{P},\mathcal{A}}\sum_{t=1}^{T}\mathbb{E}\left[ C(\bs_{t},\ba_{t}) \right] + \beta I(\bs_{1:T} \rightarrow \bz_{1:T})$};

\node (T) at (0,0) [rectangle,draw] {$T(\bs_{t+1}|\bs_{t},\ba_{t})$};

\node (P) at (-2.25,-1.9) [rectangle,draw] {$\mathcal{P}(\bz_{t}|\bs_{t},b_{t})$};
\node (P_label) at (-2.25,-2.5) {Perception unit};

\node[inner sep=0.5mm,rectangle,draw] (A) at (2.25,-1.9)  {\hspace*{-0.6cm} \begin{tabular}{ll}
     & $b_{t}=\text{predict}(\hat{b}_{t-1},\ba_{t-1})$ \\
     & $\hat{b}_{t}=\text{update}(b_{t},\bz_{t})$ \\
     & $\mathcal{A}(\ba_{t}|\hat{b}_{t})$
\end{tabular}};

\node[align=left] (A_label) at (2.25,-3) {Action unit};

\draw[->] (P) -- node[above] {$\bz_{t}$} (A);
\draw[->] (A) |- node[right] {$\ba_{t}$} (T);
\draw[->] (T) -| node[left] {$\bs_{t+1}$} (P);
\draw[->] ([xshift=-10pt]A.north) -- +(0,0.10) -| node[above,pos=0.10] {$b_{t}$} ([xshift=10pt]P.north);

\end{tikzpicture}}
        \caption{Simultaneous perception-action (SPADE) framework}
        \label{fig:Perception_diagram}
    \end{subfigure}
  \caption{Visualization of the differences between the standard POMDP framework and the SPADE framework.}
  \label{fig:POMDP_Perception_diagrams}
\end{figure}

Indeed, the proposed SPADE framework allows more flexibility for an agent compared to a conventional partially observable MDP (POMDP) (\cite{kaelbling1998planning}), with Figure \ref{fig:POMDP_Perception_diagrams} highlighting three of the key distinctions. First and foremost, our formulation includes the perception strategy $\mathcal{P}$ as a decision variable, rather than using an observation mechanism fixed a priori. 
In effect, the agent is capable of \textit{choosing} what to observe, rather than drawing an observation from a fixed sensor.
Furthermore, we allow for a belief-dependent perception mechanism, as shown in the lower left block in Figure \ref{fig:Perception_diagram}. This generalization is both biologically plausible, e.g., eye movement can be controlled, and is crucial in developing a computationally tractable synthesis of optimal perception and action strategies. Finally, in order for such a generalization to be meaningful, the SPADE framework incorporates a perception cost for the agent, denoted $I(\bs_{1:T}$$ \rightarrow $$ \bz_{1:T})$ in Fig. \ref{fig:Perception_diagram}. As we discuss in Section \ref{sec:info_theor_percep_cost}, this quantity is the directed information, a statistical measure of information flow, between the state sequence $\{\bs_{1},\ldots \bs_{t}\}$ and the observation sequence $\{\bz_{1},\ldots,\bz_{t}\}$. The inclusion of this term is critical, since full-state observation is always optimal in the absence of a perception cost.

Mathematically, our problem of interest is closest to that of stochastic optimal control with directed information constraints, as studied in \cite{tanaka2017lqg} and \cite{tanaka2021transfer}. In \cite{tanaka2017lqg}, the problem is studied in the linear-quadratic-Gaussian (LQG) regime, where it is shown that an optimal policy is comprised of linear perception and action units, whose combined structure is similar to that of Fig.~\ref{fig:Perception_diagram}. In the LQG case, the joint perception-action synthesis problem can be reformulated as a computationally-efficient semidefinite programming problem. Likewise, the finite-state counterpart of the problem was studied in \cite{tanaka2021transfer}, 
where an alternative solution method based on the so-called forward-backward Arimoto-Blahut algorithm was proposed. However, the synthesized policy does not admit the perception-action separation structure, and the algorithm suffers due to the nonconvexity of the cost function. The SPADE framework that we develop is in part motivated to overcome these difficulties.

We first show that the SPADE problem is solvable, in principle, through a backward-in-time dynamic programming algorithm over the belief space. In each iteration, each belief state value function is updated by solving a nonconvex optimization problem. However, such an approach is impractical, as the value functions must be evaluated over the entirety of the continuous belief space.
To circumvent this difficulty, we propose a novel \textit{method of invariant, finite belief sets} for approximating the value functions.
In this method, we enforce that the agent operates exclusively on an invariant, finite subset of the belief continuum. We do so by restricting the space of admissible perception strategies in such a way that the subsequent belief state always belongs to the invariant, finite subset. We show this restriction is equivalent to imposing a set of linear constraints on the set of admissible perception strategies, and that each value function update on the invariant, finite belief set can be obtained through the solution of a linear program, allowing the synthesis of an optimal simultaneous perception-action strategy through a computationally tractable and parallelizable value iteration. 

The idea of value function updates over sampled belief states is reminiscent of point-based value iteration for POMDPs (\cite{pineau2003point}). Our approach is fundamentally different since, by construction, the synthesized perception strategy renders a user-specified set of sampled belief states invariant over repeated updates, allowing for \textit{exact} value iteration over this set. For general POMDPs with observation kernels fixed a priori, exact value iteration is not possible.

We show that as the cardinality of the invariant, finite belief set increases, the linear constraints on the sensing strategies become less binding, yielding better approximations of the continuous belief space solution. Finally, as the sample density of the invariant, finite belief set increases, we prove that the value functions of the sampled belief states converge to their continuous state-space counterparts, under an assumption on the structure of these value functions.


\section{Notation}
We denote the set of all real numbers by $\mathbb{R}$. 
We use uppercase symbols to denote sets and lowercase, bold symbols to denote random variables. For a  set $X$, we denote the set of all probability distributions over $X$ by $\Delta(X)$. The probability of the event that a random variable $\bx$ takes a value $x \in X$ is denoted by $\text{Pr}(\bx=x)$. We denote a sequence of length $k$, i.e., $\{x_{1},\ldots,x_{k}\}$ by $x_{1:k}$. For a vector $V$$\in$$\mathbb{R}^{n}$, let $\text{diag}(V)$$\in$$\mathbb{R}^{n\times n}$ be a diagonal matrix of the elements of $V$.

\section{Simultaneous Perception-Action Design}

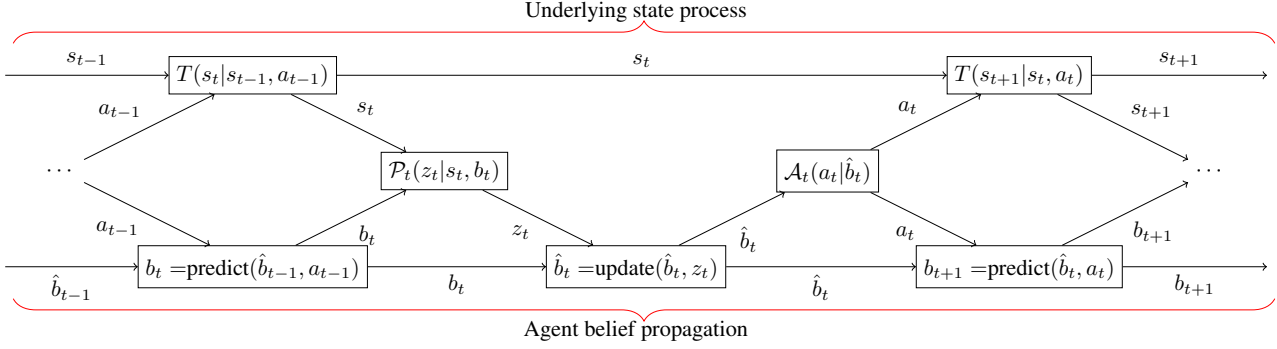
\begin{figure*}
    \centering
    \begin{tikzpicture}
    
    
    
    \node (1) at (-4,3) {};
    \node (2) at (-4,0) {};
    \node (3) at (16,3) {};
    \node (4) at (16,0) {};
    
    \node (5) at (-3,1.5) {$\cdots$};
    \node (6) at (15,1.5) {$\cdots$};
    
    \node (T1) at (0,3) [rectangle,draw] {$T(s_{t}|s_{t-1},a_{t-1})$};
    \node (T2) at (12,3) [rectangle,draw] {$T(s_{t+1}|s_{t},a_{t})$};
    \node (P) at (3,1.5) [rectangle,draw] {$\mathcal{P}_{t}(z_{t}|s_{t},b_{t})$};
    \node (A) at (9,1.5) [rectangle,draw] {$\mathcal{A}_{t}(a_{t}|\hat{b}_{t})$};
    \node (Pre1) at (0,0) [rectangle,draw] {$b_{t}=$predict$(\hat{b}_{t-1},a_{t-1})$};
    \node (Up1) at (6,0) [rectangle,draw] {$\hat{b}_{t}=$update$(\hat{b}_{t},z_{t})$};
    \node (Pre2) at (12,0) [rectangle,draw] {$b_{t+1}=$predict$(\hat{b}_{t},a_{t})$};
    
    \draw[->] (1) -- node[above] {\textcolor{black}{$s_{t-1}$}} (T1);
    \draw[->] (T1) -- node[above] {\textcolor{black}{$s_{t}$}} (T2);
    \draw[->] (T2) -- node[above] {\textcolor{black}{$s_{t+1}$}} (3);
    \draw[->] (5) -- node[above left] {\textcolor{black}{$a_{t-1}$}} (T1);
    \draw[->] (5) -- node[below left] {\textcolor{black}{$a_{t-1}$}} (Pre1);
    \draw[->] (2) -- node[below] {\textcolor{black}{$\hat{b}_{t-1}$}} (Pre1);
    \draw[->] (T1) -- node[above right] {\textcolor{black}{$s_{t}$}} (P);
    \draw[->] (Pre1) -- node[below right] {\textcolor{black}{$b_{t}$}} (P);
    \draw[->] (Pre1) -- node[below] {\textcolor{black}{$b_{t}$}} (Up1);
    \draw[->] (P) -- node[below left] {\textcolor{black}{$z_{t}$}} (Up1);
    \draw[->] (Up1) -- node[below right] {\textcolor{black}{$\hat{b}_{t}$}} (A);
    \draw[->] (A) -- node[above left] {\textcolor{black}{$a_{t}$}} (T2);
    \draw[->] (A) -- node[below left] {\textcolor{black}{$a_{t}$}} (Pre2);
    \draw[->] (T2) -- node[above right] {\textcolor{black}{$s_{t+1}$}} (6);
    \draw[->] (Pre2) -- node[below right] {\textcolor{black}{$b_{t+1}$}} (6);
    \draw[->] (Pre2) -- node[below] {\textcolor{black}{$b_{t+1}$}} (4);
    \draw[->] (Up1) -- node[below] {\textcolor{black}{$\hat{b}_{t}$}} (Pre2);
    
    \node (state_pr) at (6.0,4.0) {Underlying state process};
    \node (belief_pr) at (6.0,-1) {Agent belief propagation};
    
    \draw [decorate,decoration={brace,amplitude=10pt},color=red] (-3.75,3.5) -- (16,3.5) node [black,midway,xshift=-0.6cm] {};
    
    \draw [decorate,decoration={brace,amplitude=10pt,mirror},color=red] (-3.75,-0.5) -- (16,-0.5) node [black,midway,xshift=-0.6cm] {};
    
\end{tikzpicture}
    \caption{Interplay between the underlying state process and the prior and posterior belief states of the agent.}
    \label{fig:BeliefPropagation}
\end{figure*}

We now formulate the simultaneous perception-action design (SPADE) problem of the agent.

\subsection{Perception Model}
We use a \textit{perception MDP} $\mathcal{M}$$=$$\langle S,A,T,C,Z,\gamma \rangle$ to model the environment of the agent, where $s$$\in$$S$ is a finite set of states, $a$$\in$$A$ is a finite set of actions, $T$$:$$S$$\times$$A$$\rightarrow $$\Delta(S)$ is a transition function mapping state-action pairs to probability distributions over successor states, $C$$:$$S$$\times$$A$$\rightarrow$$\mathbb{R}$$\geq$$0$ is a cost function, $z$$\in$$Z$ is a set of observations, and $0$$\leq$$\gamma$$<$$1$ is a discount factor.
Following the formulation of \cite{shafieepoorfard2016rationally}, we assume that the agent has access to an observation space $Z$ with infinite cardinality. The rationale behind this assumption will be discussed at the end of this subsection.
We refer to the probability of transitioning to state $s'$ after taking action $a$ in state $s$ by $T(s'|a,s)$.
Likewise, we refer to the cost of taking action $a$ in state $s$ by $C(s,a)$ and the entire cost matrix by $C\in \mathbb{R}^{|S|\times |A|}$, where the rows of $C$ denote the cost of taking any action $a$$\in$$A$ for a given state $s$ while the columns denote the cost of taking action $a$ while in any state $s$$\in$$S$.

Due to imperfect information, an agent must estimate its current state $s_t\in S$ through its history of observations. Specifically, the agent maintains a \textit{belief state} $b_{t}$$=$$[b_{t}(s)$$:$$s$$\in$$S]^{\top}$$\in$$\Delta(S)$ at each time step $t$, where, for all $s$$\in$$S$, $b_{t}(s)$$=$$\text{Pr}(\bs_{t}$$=$$s|z_{1:t-1})$ denotes the probability that the agent believes it resides in state $s$ at time $t$ given the sequence of observations $z_{1:t-1}$. In our formulation, the agent maintains a parallel set of belief states over the course of its operation. We refer to the first of these belief states as the \textit{prior} belief state $b_{t}$, as defined previously. Particularly, the prior belief state characterizes the belief of the agent about its underlying state \textit{prior} to making an observation at time step $t$. Similarly, the second belief state that the agent maintains is referred to as the \textit{posterior} belief state $\hat{b}_{t}$$=$$[\hat{b}_{t}(s)$$:$$s$$\in$$S]^{\top}$$\in$$\Delta(S)$, where each $\hat{b}_{t}(s)$ is defined according to $\hat{b}_{t}(s)$$=$$\text{Pr}(\bs_{t}$$=$$s|z_{1:t})$, which is the belief state of the agent \textit{after} making an observation about its underlying state $s_{t}$ (but, we stress, before taking an action at this time step). Thus, we see that the prior belief state captures the information that the agent has available when choosing its perception strategy, while the posterior belief state captures the information that the agent has available when choosing an action. The differences between these parallel belief states are formally discussed in Section \ref{sec:RelatePriorAndPost}. 

As discussed in the introduction, in our formulation, the agent is free to design \textit{both} its action-selection strategy as well as its belief-dependent observation function. We refer to this joint process as a \textit{simultaneous perception-action strategy}, which consists of both its action strategy $\mathcal{A}$ and its perception strategy $\mathcal{P}$. To start with, the action strategy of the agent is a sequence $\mathcal{A}$$=$$\{ \mathcal{A}_{1},\ldots,\mathcal{A}_{t},\ldots\}$, where each $\mathcal{A}_{t}$$:$$\Delta(S)$$\rightarrow$$\Delta(A)$. In words, an action strategy maps a posterior belief state $\hat{b}$ at time $t$ to a probability distribution over action selection. We denote the probability of taking action $a$ in the posterior belief state $\hat{b}_{t}$ at time $t$ as $\mathcal{A}_{t}(a|\hat{b}_{t})$. Likewise, the perception strategy is a sequence $\mathcal{P}$$=$$ \{\mathcal{P}_{1},\ldots,\mathcal{P}_{t},\ldots\}$, where $\mathcal{P}_{t}$$:$$\Delta(S)$$ \times$$S$$\rightarrow$$\Delta(Z)$. The perception strategy prescribes a belief-dependent observation function for the agent at each time step. We denote the probability of making observation $z$ about the underlying state $s$ while in the prior belief state $b_{t}$ at time $t$ as $\mathcal{P}_{t}(z|s,b_{t})$.

\begin{remark}\label{rem:cardinality}
Since the agent is able to synthesize its own perception strategy $\mathcal{P}_{t}(z|s,b_{t})$, our assumption that the set $Z$ has infinite cardinality provides the agent with the greatest freedom in this synthesis problem. 
However, as shown in \cite{shafieepoorfard2016rationally}, there is no advantage to using a $Z$ with cardinality greater than that of $\Delta(S)$. In the main problem \eqref{eq:MainObjective} that we formulate below, we assume $Z=\Delta(S)$ without loss of generality, and assign each observation to a unique belief state.

\end{remark}

\subsection{Relation Between Prior and Posterior Belief States}\label{sec:RelatePriorAndPost}

Fig. \ref{fig:BeliefPropagation} details the relation between the prior and posterior belief states. Given a posterior belief state $\hat{b}_{t}$, the agent first selects an action policy $\mathcal{A}_{t}(a|\hat{b}_{t})$. Once an action policy is selected, the agent can then \textit{predict} the unique prior belief state that it transitions to by computing
\begin{equation}\label{eq:predictEquation}
    b_{t+1}(s) = \sum\nolimits_{s'\in S} \sum\nolimits_{a\in A} T(s|a,s') \mathcal{A}_{t}(a|\hat{b}_{t}) \hat{b}_{t}(s').
\end{equation}
When the action strategy is deterministic, i.e., there exists an action $a$$\in$$A$ such that $\mathcal{A}_t(a|\hat{b}_t)$$=$$1$, we may write $a$$=$$\mathcal{A}_t(\hat{b}_t)$ and more succinctly express \eqref{eq:predictEquation} as
\begin{equation}\label{eq:predictEquation2}
    b_{t+1}(s) = \sum\nolimits_{s'\in S} T(s|\mathcal{A}_t(\hat{b}_t),s')  \hat{b}_{t}(s').
\end{equation}
Once the agent has transitioned to the prior belief state $b_{t}$, it then chooses its perception strategy $\mathcal{P}_{t}(z|s,b_{t})$. With probability $\text{Pr}(\bz_{t}$$=$$z)$$=$$\sum\nolimits_{s \in S}\mathcal{P}_{t}(z|s,b_{t})b_{t}(s)$, the agent then makes an observation $z_{t}$$=$$z$ and \textit{updates} to the posterior belief state $\hat{b}_{t}$ according to the set of Bayesian updates
\begin{equation}\label{eq:updateEquation}
    \hat{b}_{t}(s) =  \frac{\mathcal{P}_{t}(z|s,b_{t})b_{t}(s)}{\sum_{s'\in S}\mathcal{P}_{t}(z|s',b_{t})b_{t}(s')}
\end{equation}
for each $s$$\in$$S$. Note that the transition to the posterior belief state $\hat{b}_{t}$ occurs with probability $\text{Pr}(\bz_{t}$$=$$z)$. Note that when $Z$ is a countable set, the structure of \eqref{eq:updateEquation} implies that, for any $b_{t}$$\in$$\Delta(S)$, $\hat{b}_{t}$ is a collection of point masses 
each having weight $\text{Pr}(\bz_{t}$$=$$z)$. 
We adopt the notation $\hat{b}_t^z$ to denote the unique posterior belief obtained when observation $z$ is made while in prior belief state $b_t$.
Similarly, we denote $b_{t+1}^{z,a}$ as the unique prior belief obtained when action $a$ is chosen in posterior belief state $\hat{b}_t^z$.

\subsection{An Information-Theoretic Perception Cost}\label{sec:info_theor_percep_cost}

In the absence of perception costs, it is always optimal for the agent to select a noiseless, full-state measurement (i.e., $Z=S$ and $z_t=s_t$) as the perception strategy. To make the perception design problem meaningful, it is necessary to introduce a mathematical metric to model the cost of information acquisition for the agent.
Although the SPADE framework is general enough to support a variety of perception cost functions, we focus on a model where these costs are quantified by the information-theoretic concept of \textit{directed information}. Directed information (\cite{massey1990causality}), sometimes referred to as transfer entropy (\cite{schreiber2000measuring}), is a non-negative quantity characterizing the information flow between random processes. For a pair of random processes $\bs_{1:T}$ and $\bz_{1:T}$, the directed information from $\bs_{1:T}$ to $\bz_{1:T}$ is 
%
\begin{equation*}
    I(\bs_{1:T}\rightarrow \bz_{1:T}) \triangleq\sum\nolimits_{t=1}^T I(\bs_{1:t};\bz_t|\bz_{1:t-1}),
\end{equation*}    
where $I(\bs_{1:t};\bz_t|\bz_{1:t-1})$ is the conditional mutual information (\cite{cover2012elements}), explicitly written as
\begin{equation*}
    \sum_{s_{1:t},z_{1:t}} \text{Pr}(s_{1:t},z_{1:t})\log \frac{\text{Pr}(s_{1:t},z_{t}|z_{1:t-1})}{\text{Pr}(s_{1:t}|z_{1:t-1})\text{Pr}(z_{t}|z_{1:t-1})}.
\end{equation*}
%
%
In the SPADE model, the random processes $\bs_{1:T}$ and $\bz_{1:T}$ represent the state and the observation sequences, respectively. 
The directed information is closely related to the information traffic from the perception unit to the action unit (the lower left and right blocks of Fig. \ref{fig:Perception_diagram}, respectively), and is a suitable metric to capture perception costs as, for our model, it is equivalent to the summation of the stage-additive information gains. A formal analysis providing the directed information with a Shannon-theoretic operational meaning is provided in Appendix~\ref{sec:appendix_perception_cost}. For mathematical convenience, we introduce the discounted directed information:
\begin{align*}
I_\gamma(\bs_{1:T}\rightarrow \bz_{1:T}) & \triangleq\sum\nolimits_{t=1}^\infty \gamma^{t-1} I(\bs_{1:t};\bz_t|\bz_{1:t-1}) \\
& =\sum\nolimits_{t=1}^\infty \gamma^{t-1} I(\bs_t;\bz_t|\bz_{1:t-1}).
\end{align*}
\subsection{Main Problem}

In the SPADE problem, the objective of the agent is to minimize the discounted sum of its perception and environmental costs. The agent accomplishes this objective by optimally choosing its perception and action strategies, which we express through the objective function
\begin{equation}\label{eq:MainObjective}
    \min_{\mathcal{P}_{t},\mathcal{A}_{t}} \sum\nolimits_{t=1}^{\infty} \gamma^{t-1} \left[ \beta I(\bs_{t};\bz_{t}|\bz_{1:t-1}) + \mathbb{E}\left[ C(\bs_{t},\ba_{t}) \right] \right],
\end{equation}
where $\beta$ is a parameter weighting the relative cost of information. We seek to formulate our objective in (\ref{eq:MainObjective}) in terms of a dynamic programming problem over the belief space $\Delta(S)$. To this end, consider a prior belief state with $b_{t}(s)$$=$$ \text{Pr}(\bs_{t}$$=$$s|z_{1:t-1})$. The stage-wise mutual information for this belief state $b_{t}$ is 
\begin{align}
    I(\bs_{t};\bz_{t}|\bz_{1:t-1}) & = \sum_{s\in S} \int_{Z} \mathcal{P}_{t}(z|s,b_{t}) b_{t}(s) \cdot \nonumber \\
    & \qquad \qquad \log \frac{\mathcal{P}_{t}(z|s,b_{t})}{\sum\limits_{s'\in S}\mathcal{P}_{t}(z|s',b_{t})b_{t}(s')} \nonumber \\
    & \triangleq R(b_{t},\mathcal{P}_{t}). \label{eq:MutInfoBetStaObs}
\end{align}
Note that for a fixed $b_{t}$, the stage-wise mutual information in (\ref{eq:MutInfoBetStaObs}) is convex with respect to the perception strategy $\mathcal{P}_{t}$, since it is the weighted sum of relative entropies. 

From the definitions of $R(b_{t},\mathcal{P}_{t})$ and $C(s_{t},a_{t})$, we can decompose the objective function in \eqref{eq:MainObjective} into a set of recursive Bellman equations, where the value function is given by
\begin{align}
    & V(b_{t}) \triangleq \inf_{\mathcal{A}_{t}, \mathcal{P}_{t}} \big\{ \beta R(b_{t},\mathcal{P}_{t}) \nonumber \\
    & \quad  + \mathbb{E}_{b_{t}}^{\mathcal{A}_{t},\mathcal{P}_{t}}\left[ C(\bs_{t},\ba_{t}) \right] + \gamma \mathbb{E}_{b_{t}}^{\mathcal{A}_{t},\mathcal{P}_{t}}\left[ V(b_{t+1}^{\bz,\ba}) \right] \big\} \label{eq:recursive_bellman}
\end{align}
for each prior belief state $b_{t}$ at time $t$. 
The notation $\mathbb{E}_{b_{t}}^{\mathcal{A}_{t},\mathcal{P}_{t}}$ emphasizes that these expectations are evaluated under the joint distribution defined by $b_t, \mathcal{P}_{t}$ and $\mathcal{A}_{t}$.
From a prior belief state $b_{t}$, recall that the agent first transitions to a posterior belief state $\hat{b}_{t}$ through the update equation (\ref{eq:updateEquation}) with probability given by $\text{Pr}(\bz_{t}$$=$$z)$. Once in $\hat{b}_{t}$, the agent then transitions to a prior belief state $b_{t+1}$ through the predict equation (\ref{eq:predictEquation}). Through this evolution, we can decompose (\ref{eq:recursive_bellman}) into the perception and action stages, respectively:
\begin{subequations}
    \begin{align}
        & V(b_{t}) = \inf_{\mathcal{P}_{t}} \huge[\beta  R(b_{t},\mathcal{P}_{t}) \nonumber \\
        & \qquad \qquad + \sum_{s\in S}\int_{z\in Z} \mathcal{P}_{t}(z|s,b_{t})b_{t}(s)\hat{V}(\hat{b}_{t}^{z})dz \huge], \label{eq:expProbsBellman}
    \end{align}
    \begin{equation}\label{eq:simplifiedPosVals}
        \hat{V}(\hat{b}_{t}^{z}) = \min_{a} \left[ \sum_{s \in S} \hat{b}_{t}^{z}(s)C(s,a) + \gamma V(b_{t+1}^{z,a}) \right],
    \end{equation}
\end{subequations}
for each $z$$\in$$Z$. In (\ref{eq:expProbsBellman}), we have explicitly written the expectation by noting that, for a given prior belief state $b_{t}$, the posterior belief state $\hat{b}_{t}^{z}$ is a random variable realized with a probability $\text{Pr}(\bz_{t}$$=$$z)$ and a state distribution given by the update equation (\ref{eq:updateEquation}). To obtain (\ref{eq:simplifiedPosVals}), recall for a posterior belief state $\hat{b}_{t}^{z}$,
each action $a$ yields a unique transition to the prior belief state $b_{t+1}^{z,a}$ given by the predict equation \eqref{eq:predictEquation2}.
In \eqref{eq:simplifiedPosVals}, $\min_{\mathcal{A}_t}$ is replaced with $\min_a$ since it is straightforward to show the optimal action strategies are deterministic.

The combined set of recursive Bellman equations given by (\ref{eq:expProbsBellman}) and (\ref{eq:simplifiedPosVals}) suggests the use of dynamic programming to solve our main problem (\ref{eq:MainObjective}). 
Namely, denote by $B(\Delta(S))$ the space of functions $V:\Delta(S)\rightarrow \mathbb{R}$ such that $\|V\|_{\text{sup}}\triangleq\sup_{b\in\Delta{S}}|V(b)|<+\infty$.
Now, define the operator $T$ by
\begin{align}\label{eq:def_operator_T}
    (TV)(b) & \triangleq \inf_{ \mathcal{A}, \mathcal{P}} \{ \beta R(b,\mathcal{P})  \nonumber \\
    & + \mathbb{E}_{b}^{\mathcal{A},\mathcal{P}}[C(\bs,\ba)]  + \gamma \mathbb{E}_{b}^{\mathcal{A},\mathcal{P}}[V(b^{\bz,\ba})] \}.
\end{align}
Using $T$, the Bellman equation \eqref{eq:recursive_bellman} can be written as $V=TV$.
The following theorem states that $T$ is a contractive mapping from $B(\Delta(S))$ to itself and that the corresponding value iteration is convergent.
\begin{restatable}{thm}{thm_T}\label{thm_T} The following results hold for the operator $T$:
\begin{itemize}
    \item[(a)] For any $V\in B(\Delta(S))$ and $V'\in B(\Delta(S))$,
    \begin{align*}
        \|TV - TV'\|_{\text{sup}} \leq \gamma \|V - V'\|_{\text{sup}}.
    \end{align*}
    \item[(b)] For an arbitrary $V_0\in B(\Delta(S))$, define a sequence of functions $\{V_k\}_{k=1,2,...}$, $V_k\in B(\Delta(S))$, by $V_k=T^k V_0$, $k=1,2, ...$. Then,  we have
    \[
    \lim_{k\rightarrow \infty} \|V_k-V^*\|_{\text{sup}}=0
    \]
    where $V^*$$\in$$B(\Delta(S))$ is the unique solution to $V^*$$=$$TV^*$.
\end{itemize}
\end{restatable}
\textbf{Proof:}
(a) Let $q\triangleq \|V - V'\|_{\text{sup}}$. Then,
\begin{align*}
    V(b) - q \leq V'(b) \leq V(b) + q
\end{align*}
for every $b\in \Delta(S)$. Applying the operator $T$ to each side of the inequality, we have that, for each $b \in \Delta(S)$,
\begin{align*}
    TV(b) - \gamma q \leq TV'(b) \leq TV(b) + \gamma q,
\end{align*}
where we have made use of the fact that 
\begin{align*}
    \gamma \mathbb{E}_{b}^{\mathcal{A}, \mathcal{P}}[V(b^{\bz,\ba}) + q] = \gamma \mathbb{E}_{b}^{\mathcal{A}, \mathcal{P}}[V(b^{\bz,\ba})] + \gamma q
\end{align*}
in \eqref{eq:def_operator_T}. The result then follows.

(b) Notice that the space $B(\Delta(S))$ equipped with the sup norm 
$\|\cdot \|_{\text{sup}}$ is a complete metric space. Since $T$ is a contractive mapping from $B(\Delta(S))$ to itself, we can apply the Banach fixed-point theorem (\cite{khamsi2011introduction}) to obtain the result.
$\Box$

The direct implementation of the value iteration $V_k=T^k V_0$ is computationally intractable as the function $V(\cdot)$ must be evaluated everywhere on the continuous belief space $\Delta(S)$.

\begin{remark}
    It is possible to express our main objective \eqref{eq:MainObjective} in terms of the standard POMDP paradigm described in Fig. \ref{fig:POMDP_diagram} by augmenting the space of actions with the space of perception strategies; i.e., $\mathcal{A}_{\text{aug}}$$=$$(\mathcal{P}, \mathcal{A})$. Suppose that we then discretize the space of perception strategies such that $\mathcal{A}_{\text{aug, disc}}$$=$$(\mathcal{P}_{\text{disc}}, \mathcal{A})$ contains only a finite number of actions. By doing so, we have converted the problem into the standard form of a POMDP, for which we can use off-the-shelf POMDP solvers. However, this approach introduces two layers of approximation. One must first approximate the continuous space $\mathcal{P}$ of perception kernels by a finite set $\mathcal{P}_{\text{disc}}$. The second approximation is due to the inherent hardness of POMDPs; standard PBVI provides a universal scheme to approximately solve a POMDP by discretizing the belief space. It is not clear how to cleverly perform both discretizations to achieve the best computational performance.
\end{remark}

\section{Method of Invariant Finite Belief Sets}

Due to the continuity of both $\Delta(S)$ and the set $Z$, exactly solving for an optimal simultaneous perception-action strategy is computationally prohibitive. We now focus on developing a method approximating (\ref{eq:expProbsBellman}) and (\ref{eq:simplifiedPosVals}). We develop a novel method in which the agent operates on a finite subset of the continuous belief space, where the chosen subset remains invariant over time. We refer to this set as an \textit{invariant finite belief set} (IFBS). We then show that, as the number of sampled belief states approaches infinity, the value functions converge to their continuous state-space counterparts, under an appropriate assumption. For notational clarity, we omit the time index $t$ of all variables.

\subsection{Method of Invariant Finite Belief Sets}

To construct a model approximating (\ref{eq:expProbsBellman}) and (\ref{eq:simplifiedPosVals}), we first pick a representative  sample of posterior belief states, $\hat{\mathcal{B}}$$\subset$$\Delta(S)$. The set $\hat{\mathcal{B}}$ consists of a finite number of elements, which can be arbitrarily
\begin{figure}
    \centering
    \includegraphics[width=0.45\textwidth]{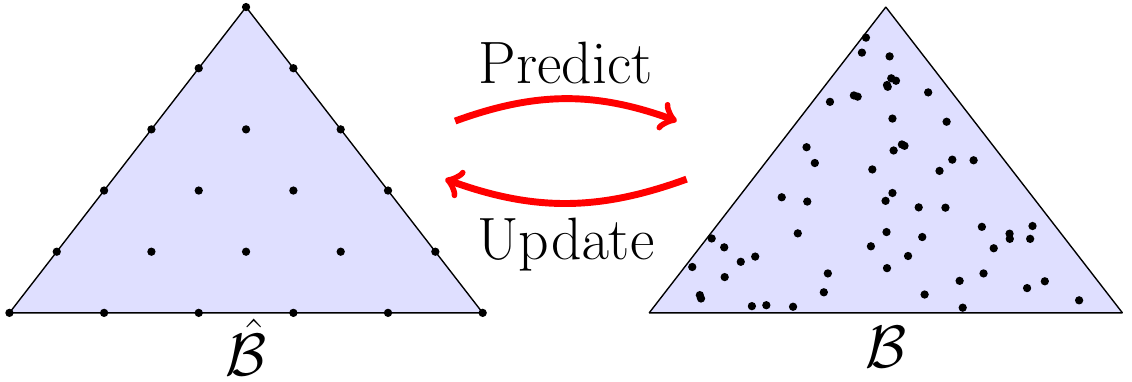}
    \caption{As a visualization, consider the set $\hat{\mathcal{B}}$, which consists of the uniformly sampled belief states (black dots) drawn from the belief simplex shown at left. In turn, the set of belief states $\mathcal{B}$ is obtained by applying \eqref{eq:predictEquation} to each belief state in $\hat{\mathcal{B}}$ for each action $a\in \mathcal{A}$. Note that the representative belief states in both $\hat{\mathcal{B}}$ and $\mathcal{B}$ shown above remain invariant over repeated predict and update steps.}
    \label{fig:SeqPosPri}
\end{figure}
chosen (e.g. a uniform gridding, as in Fig. \ref{fig:SeqPosPri}) as long as the following assumption is met.
\begin{assumption}\label{assumption1_EP}
    The set $\hat{\mathcal{B}}$ contains all extreme points of the belief simplex.
\end{assumption}
%
%
Recall from Remark~\ref{rem:cardinality} that we set $Z=\Delta(S)$ without loss of generality for the continuous state-space problem. Similarly, for the method of invariant finite belief sets, we set $Z=\hat{\mathcal{B}}$ without loss of generality. Thus, we have that $|Z|=|\hat{\mathcal{B}}|=M$. Based on the selection of $\hat{\mathcal{B}}$, we can apply the update equation (\ref{eq:updateEquation}) for each $\hat{b}$$\in$$\hat{\mathcal{B}}$ and each action $a$$\in$$A$ to obtain a corresponding finite set of prior belief states, which we refer to as $\mathcal{B}$$\subset$$\Delta(S)$. Furthermore, since
the subsequent prior belief state $b$ given a posterior belief state $\hat{b}$ and action $a$ is unique, we have that $|\mathcal{B}|$$=$$M$$\cdot$$|A|$. 

%
Given the sets $\mathcal{B}$ and $\hat{\mathcal{B}}$, we now seek conditions under which they are \textit{invariant}; i.e., over repeated predict and update steps using (\ref{eq:predictEquation}) and (\ref{eq:updateEquation}), the agent should operate exclusively within $\mathcal{B}$ and $\hat{\mathcal{B}}$, as illustrated in Fig. \ref{fig:SeqPosPri}. By the construction of the set $\mathcal{B}$, the predict equation (\ref{eq:predictEquation}) trivially yields prior belief states exclusively on $\mathcal{B}$ for an agent in any $\hat{b}$$\in$$\hat{\mathcal{B}}$ taking any action. It remains to show that we can restrict the update equation such that the resulting posterior belief state remains in the set $\hat{\mathcal{B}}$.
This condition requires that for each prior belief state $b$$\in$$\mathcal{B}$ and for any observation $z$$\in$$Z$ made while in $b$, the resulting posterior belief state is guaranteed to exist in the set $\hat{\mathcal{B}}$; i.e., $\hat{b}^{z}$$\in$$\hat{\mathcal{B}}$, where, as previously discussed, $\hat{b}^{z}$ is the posterior belief state that results from making observation $z$ while in prior belief state $b$.
Recalling the update equation (\ref{eq:updateEquation}), the individual probabilities $\hat{b}^{z}(s)$ are
\begin{equation}\label{eq:probUpdate}
    \hat{b}^{z}(s) =  \frac{\mathcal{P}(z|s,b)b(s)}{\sum_{s'\in S}\mathcal{P}(z|s',b)b(s')} \quad \forall s \in S.
\end{equation}
To ensure that the posterior belief state $\hat{b}^{z}$ obtained by 
\eqref{eq:probUpdate} remains in our invariant set $\hat{\mathcal{B}}$,
we seek to impose restrictions on the set of admissible perception strategies $\mathcal{P}$ such that $\hat{b}^z\in\hat{\mathcal{B}}$ is guaranteed for all possible observations $z\in Z$.
Recalling Remark \ref{rem:cardinality}, such restrictions can readily be imposed since the sets $\hat{\mathcal{B}}$ and $Z$ have equal cardinalities, as well as the fact that the perception strategy is belief-dependent. 

In what follows, we will show that these restrictions are linear constraints on $\mathcal{P}$, and that they are algorithmically straightforward to incorporate. To start with, recall that the prior and posterior belief states are vectors given by $b$$=$$[b(s)$$:$$s$$\in$$S]^{\top}$ and $\hat{b}$$=$$[\hat{b}(s)$$:$$s$$\in$$S]^{\top}$, respectively. We now introduce $\mathcal{P}^{z}_{b}$$\in$$\mathbb{R}^{|S|}$ for each observation $z$$\in$$Z$ and prior belief state $b$$\in$$\mathcal{B}$ to denote the vector $[\mathcal{P}(z|s,b)$$:$$s$$\in$$S]^{\top}$$\in$$ \mathbb{R}^{|S|}$. Using this notation, we subsequently introduce the scalar
\begin{equation*}
\alpha^{z}_{b}\triangleq\sum\nolimits_{s \in S} \mathcal{P}(z|s,b) b(s)=b^{\top} \mathcal{P}^{z}_{b}
\end{equation*}
to encode the probability of observing $z$ when in the prior belief state $b$ and the perception strategy $\mathcal{P}$ is applied. Equation (\ref{eq:probUpdate}) implies $\alpha^{z}_{b}$ and $\mathcal{P}_{b}^{z}$ have the linear relation
\begin{equation}\label{eq:vectorForm}
    \alpha^{z}_{b} \, \hat{b}^{z} = \text{diag}(b) \mathcal{P}^{z}_{b}.
\end{equation}
We can use this constraint to ensure that the updated posterior belief state remains in the set $\hat{\mathcal{B}}$ as follows. For a fixed prior belief state $b$, since the cardinalities of both the set $\hat{\mathcal{B}}$ and $Z$ are equal, we can assign each $z$ in \eqref{eq:probUpdate} to a unique posterior belief state $\hat{b}$$\in$$\hat{\mathcal{B}}$. Explicitly writing the set $\hat{\mathcal{B}}$ as $\hat{\mathcal{B}}$$=$$\{ \hat{b}^{1},\ldots,\hat{b}^{M} \}$, and the set $Z$ as $Z$$=$$\{ 1, \ldots, M\}$, we can alternatively express \eqref{eq:vectorForm} as
\begin{equation}\label{eq:vectorFormExplicit}
    \alpha^m_{b} \, \hat{b}^{m} = \text{diag}(b) \mathcal{P}^m_{b} \quad \forall m=1\ldots M,
\end{equation}
where we have assigned, without loss of generality, the $m^{th}$ observation to update the prior belief state $b$ to the $m^{th}$ posterior belief state $b^{m}$ through \eqref{eq:probUpdate}.
Rearranging terms in (\ref{eq:vectorFormExplicit}) and writing it for each $m$$=$$1,\ldots,M$, we obtain
\begin{equation}\label{eq:linearConstraints}
    \setlength\arraycolsep{2pt}
    \left[\begin{array}{c;{4pt/4pt}c}
        \begin{matrix} \text{diag}(b) & { } & 0 \\ { } & \ddots & { } \\ 0 & { } & \text{diag}(b) \end{matrix} & \begin{matrix} -\hat{b}^{1} & { } & 0 \\ { } & \ddots & { } \\ 0 & { } & -\hat{b}^{M} \end{matrix} \\ \hdashline[4pt/4pt]
        \begin{matrix} I_{|S|} & \cdots & I_{|S|} \end{matrix} & 0_{|S|\times M}
    \end{array}\right]
    \left[\begin{array}{c}
        \mathcal{P}^1_{b} \\
        \vdots \\
        \mathcal{P}^{M}_{b} \\
        \hdashline
        \alpha^{1}_{b} \\
        \vdots \\
        \alpha^{M}_{b}
    \end{array}\right] = 
    \left[\begin{array}{c}
        0 \\
        \vdots \\
        0 \\
        \hdashline
        1 \\ 
        \vdots \\
        1
    \end{array}\right],
\end{equation}
%
%
%
where the lower set of constraints additionally encodes that the perception strategy must be a valid probability distribution; i.e., $\sum_{z\in Z} \mathcal{P}(z|s,b) = 1$ for each state $s\in S$.
Since valid probability distributions must have nonnegative elements, we finally impose that
\begin{equation}\label{eq:entry_wise}
    \left[
        \mathcal{P}^{z}_{b},
        \alpha^{z}_{b}
    \right]^{\top} \geq 0, \quad \forall z\in Z
\end{equation}
entry-wise. Assuming that the number of observations is greater than the number of states; i.e., $M$$>$$|S|$, the set of constraints in (\ref{eq:linearConstraints}) is an underdetermined linear system with $(M$$\cdot$$|S|$$+$$M)$ variables and $(M$$\cdot$$|S|$$+$$|S|)$ constraints, so there exist infinitely many solutions to (\ref{eq:linearConstraints})-(\ref{eq:entry_wise}). We note that in realistic scenarios, we typically have that $M$$\gg$$|S|$. We now state the following lemma.
\begin{restatable}{lemma}{feassolution}\label{Lemma1}
    Under Assumption \ref{assumption1_EP}, (\ref{eq:linearConstraints})-(\ref{eq:entry_wise}) admits a feasible solution.
\end{restatable}
\noindent
\textbf{Proof:} See Appendix~\ref{sec:theory_proofs}. $\Box$

In what follows, we denote $\mathcal{P}(b$$\rightarrow$$\hat{\mathcal{B}})$ as the subset of $\mathcal{P}$ satisfying the linear constraints (\ref{eq:linearConstraints})-(\ref{eq:entry_wise}).

\subsection{Dynamic Programming Revisited}

We now propose a method to approximate the dynamic programming formulas (\ref{eq:expProbsBellman}) and (\ref{eq:simplifiedPosVals}) using our invariant, finite belief set. For the user-defined set of posterior belief states $\hat{\mathcal{B}}$ satisfying Assumption 1, denote its associated set of prior belief states as $\mathcal{B}$. For each $b$$\in$$\mathcal{B}$, we modify (\ref{eq:expProbsBellman}) to
\begin{align}\label{eq:modifiedBellman}
    & V(b) = \min_{\mathcal{P}\in \mathcal{P}(b\rightarrow\hat{\mathcal{B}})} \Huge[ \beta R(b,\mathcal{P}) \nonumber \\
    & \qquad \qquad + \sum\nolimits_{\substack{s\in S, \\ z\in Z}} \mathcal{P}(z|s,b)b(s)\hat{V}(\hat{b}^{z}) \Huge].
\end{align}
We have modified (\ref{eq:expProbsBellman}) by including an additional constraint $\mathcal{P}$$\in$$\mathcal{P}(b$$\rightarrow$$\hat{\mathcal{B}})$ which ensures that the perception strategy causes the agent to remain on the IFBS. Notably, (\ref{eq:modifiedBellman}) is a convex optimization problem and can be further simplified. In the following lemma, we show that (\ref{eq:modifiedBellman}) can be reduced to an equivalent linear program (LP).
\begin{restatable}{lemma}{linearprogram}\label{Lemma3}
    For a given prior belief state $b$, define
    \begin{align*}
        \overline{S}(b) & \triangleq \{ s \, | \, b(s) \neq 0 \} \\
        \overline{M}(b) & \triangleq \{ m \, | \, \mathrm{supp}(\hat{b}^{m}) \subseteq \mathrm{supp}(b) \}
    \end{align*}
    Now, introduce the notation $b[\overline{S}(b)]$$\triangleq$$\text{col}_{s\in\overline{S}(b)}\{b(s)\}$ and  $\hat{b}[\overline{S}(b)]$$\triangleq$$\text{col}_{s\in\overline{S}(b)}\{\hat{b}(s)\}$. Then, (\ref{eq:modifiedBellman}) is equivalent to the following LP:
    %
    \begin{subequations}
    \label{eq:lin_prog-0}
        \begin{align}
            \min_{\alpha_{b}\geq 0} & \, \, \sum\nolimits_{m\in \overline{M}(b)} F_{m} \alpha_{b}^{m} \label{eq:lin_prog-1} \\
            \mathrm{ s.t.} & \,\, \sum\nolimits_{m\in \overline{M}(b)} \alpha_{b}^{m} \hat{b}^{m}[\overline{S}(b)] = b[\overline{S}(b)] \label{eq:lin_prog-2} \\
            & \, \, \alpha_{b}^{m} = 0 \quad \forall m \not\in \overline{M}(b),\label{eq:lin_prog-3}
        \end{align}
    \end{subequations}
    where $\alpha_{b} = [\alpha^{1}_{b},\ldots,\alpha^{M}_{b}]^{\top}\in\mathbb{R}^M$ is the decision variable and
    \begin{align*}
        & F_{m} = \beta D(\hat{b}^{m}||b) + \hat{V}(\hat{b}^{m}), \\
        & D(\hat{b}^{m}||b) = \sum\nolimits_{s\in \overline{S}(b)}\hat{b}^{m}(s) \log \frac{\hat{b}^{m}(s)}{b(s)}.
    \end{align*}
\end{restatable}
\noindent
\textbf{Proof:} See Appendix~\ref{sec:theory_proofs}. $\Box$

In brief, the proof of Lemma \ref{Lemma3} proceeds as follows. From (\ref{eq:vectorForm}), by removing states with $b(s)$$=$$0$, we can invert $\text{diag}(b)$ to obtain an explicit parameterization of $\mathcal{P}_{b}^{z}[\overline{S}(b)]$ in terms of $\alpha^{z}_{b}$. We can then substitute this relation for each individual perception strategy variable into (\ref{eq:modifiedBellman}). Substituting for the objective function yields~\eqref{eq:lin_prog-1}, while substituting into the second set of constraints in (\ref{eq:linearConstraints}) yields the given equality constraints, producing the desired LP.




We introduce $C_{a}$$\in$$ \mathbb{R}^{|S|}$ to denote the column of the cost matrix $C$ corresponding to a given action $a$$\in$$A$. 
Then, for each $\hat{b}^m\in \mathcal{\hat{B}}$, we can express (\ref{eq:simplifiedPosVals}) as
\begin{equation}\label{eq:modModPosBel}
    \hat{V}(\hat{b}^m) = \min_{a\in A} \left[ C_{a}^{\top}\hat{b}^m + \gamma V(b^{m,a}) \right].
\end{equation}
For a given IFBS, both \eqref{eq:lin_prog-0} and (\ref{eq:modModPosBel}) are computationally tractable. Furthermore, \eqref{eq:lin_prog-0} is parallelizable for each $b_{t}$$\in$$\mathcal{B}$. Based on this discussion, the following backward dynamic programming problem is suggested: for each $b$$\in$$ \mathcal{B}$, solve \eqref{eq:lin_prog-0}, while 
for each $\hat{b}$$\in$$\hat{\mathcal{B}}$, solve (\ref{eq:modModPosBel}).

Applying \eqref{eq:modModPosBel} to the vector $V$ followed by the operation \eqref{eq:lin_prog-0} is equivalent to applying the operator $\tilde{T}: \mathbb{R}^M \rightarrow \mathbb{R}^M$ defined according to
\begin{align}\label{eq:def_operator_T_tilde}
    (\tilde{T}\tilde{V})(b) & = \min_{\mathcal{A}, \mathcal{P}\in \mathcal{P}(b\rightarrow\hat{\mathcal{B}})} \{ \beta R(b,\mathcal{P})  \nonumber \\
    & + \mathbb{E}_{b}^{\mathcal{A},\mathcal{P}}[C(\bs,\ba)] + \gamma \mathbb{E}_{b}^{\mathcal{A},\mathcal{P}}[\tilde{V}(b^{\bm,\ba})] \}.
\end{align}
For an initial bounded vector $\tilde{V}_0\in\mathbb{R}^M$, the value iteration procedure $\tilde{V}_k=\tilde{T}^k\tilde{V}_0$ can be viewed as an approximation of the original value iteration $V_k=T^kV_0$ in Theorem~\ref{thm_T}.
Similar to Theorem~\ref{thm_T}, the convergence of the modified value iteration procedure is readily shown. This result is formalized in Theorem~\ref{thm_T_tilde} below.
\begin{restatable}{thm}{thm_T_tilde}\label{thm_T_tilde}
The following results hold for the operator $\tilde{T}$:
\begin{itemize}
    \item[(a)] For any bounded vectors $\tilde{V}\in \mathbb{R}^M$ and $\tilde{V}'\in \mathbb{R}^M$,
    \begin{align*}
        \|\tilde{T}\tilde{V} - \tilde{T}\tilde{V}'\|_\infty \leq \gamma \|\tilde{V} - \tilde{V}'\|_\infty.
    \end{align*}
    \item[(b)] For an arbitrary bounded vector $\tilde{V}_0\in \mathbb{R}^M$, define a sequence of bounded vectors $\{\tilde{V}_k\}_{k=1,2,...}$ by $\tilde{V}_k=\tilde{T}^k \tilde{V}_0$, $k=1,2, ...$. Then, we have
    \[
    \lim_{k\rightarrow \infty} \tilde{V}_k=\tilde{V}^*
    \]
    where $\tilde{V}^*\in \mathbb{R}^M$ is the unique solution to $\tilde{V}^*=\tilde{T}\tilde{V}^*$.
\end{itemize}
\end{restatable}
\textbf{Proof:}
The proof is similar to that of Theorem~\ref{thm_T} and hence is omitted.
$\Box$

To limit computation time, it is advantageous to construct $\hat{\mathcal{B}}$ using as few representative belief states as possible. However, doing so may yield value functions in (\ref{eq:lin_prog-1})-(\ref{eq:lin_prog-3}) and (\ref{eq:modModPosBel}) that poorly approximate the true belief space value functions of (\ref{eq:expProbsBellman}) and (\ref{eq:simplifiedPosVals}). Thus, it is desirable to possess some method for improving the approximation. To this end, we present the following Lemma.
\begin{restatable}{lemma}{monotonicity}\label{lem:monotonicity}
    Let $\hat{\mathcal{B}}$ be a set of posterior belief states with the corresponding set of prior belief states $\mathcal{B}$ found through the solution of (\ref{eq:predictEquation}) for each $\hat{b}$$\in$$\hat{\mathcal{B}}$ and $a$$\in$$A$. Denote their respective value functions by $\hat{V}(\cdot)$ and $V(\cdot)$. Consider a new set $\hat{\mathcal{B}}'$$=$$\hat{\mathcal{B}}$$\cup$$\hat{b}$ where $\hat{b} \not\in \hat{\mathcal{B}}$; i.e., $\hat{\mathcal{B}}'$ is formed by adding a sample belief state $\hat{b}$ to $\hat{\mathcal{B}}$. For $\hat{\mathcal{B}}'$, denote its associated set of prior belief states by $\mathcal{B}'$ and their value functions by $\hat{V}'(\cdot)$ and $V'(\cdot)$, respectively. Then, $V(b)$$\geq$$V'(b)$ for all $b$$\in$$\mathcal{B}$ and $\hat{V}(\hat{b})$$\geq$$\hat{V}'(\hat{b})$ for all $\hat{b}$$\in$$ \hat{\mathcal{B}}$; i.e., the value functions are monotonically non-increasing as $|\hat{\mathcal{B}}|$ increases.
\end{restatable}
\noindent
\textbf{Proof:} See Appendix~\ref{sec:theory_proofs}. $\Box$
\begin{figure}
    \centering
    \begin{subfigure}[b]{0.45\textwidth}
        \centering\scalebox{0.45}{
        \includegraphics[]{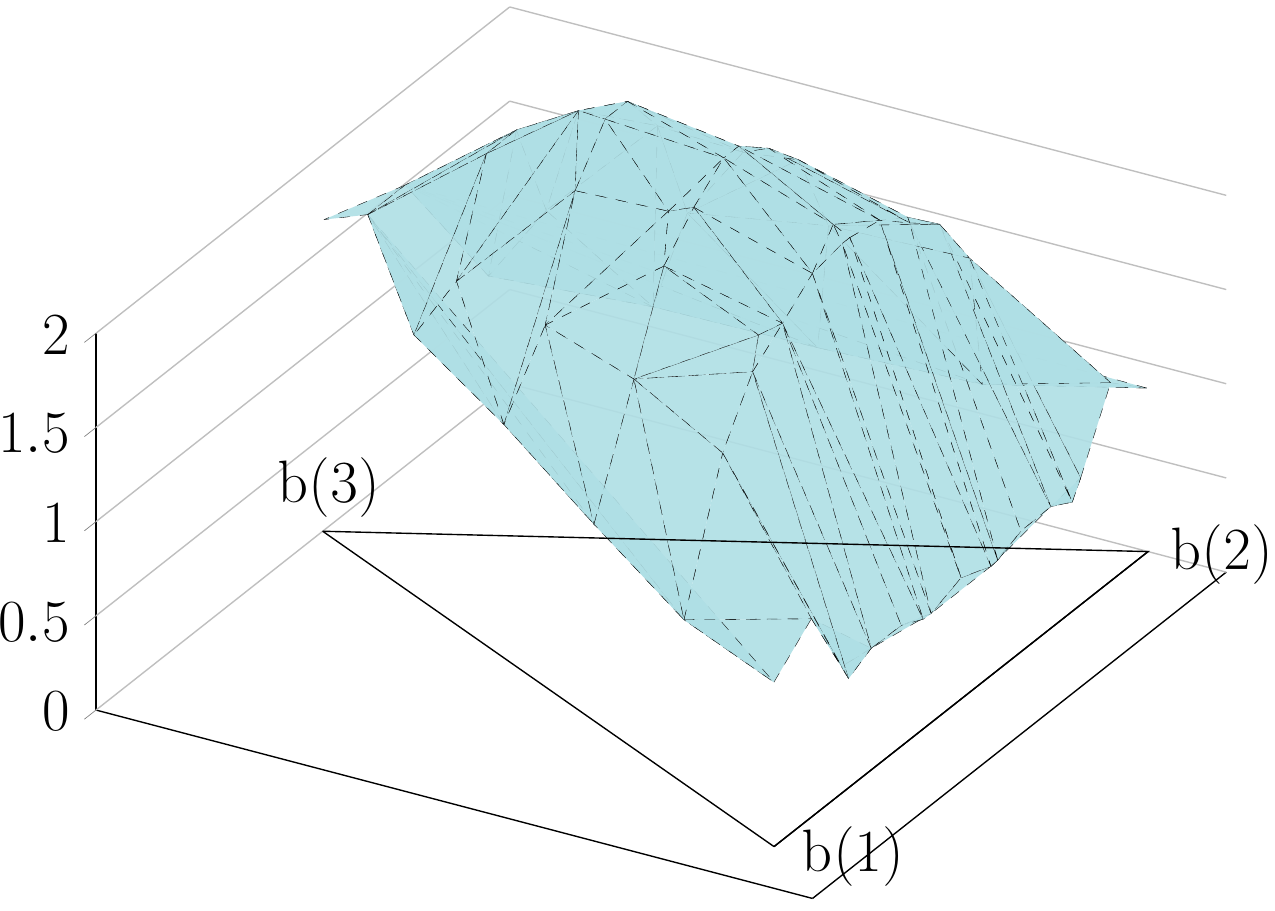}}
        \label{fig:increase_samples_coarse}
    \end{subfigure}
    \begin{subfigure}[b]{0.45\textwidth}
        \centering\scalebox{0.45}{
        \includegraphics[]{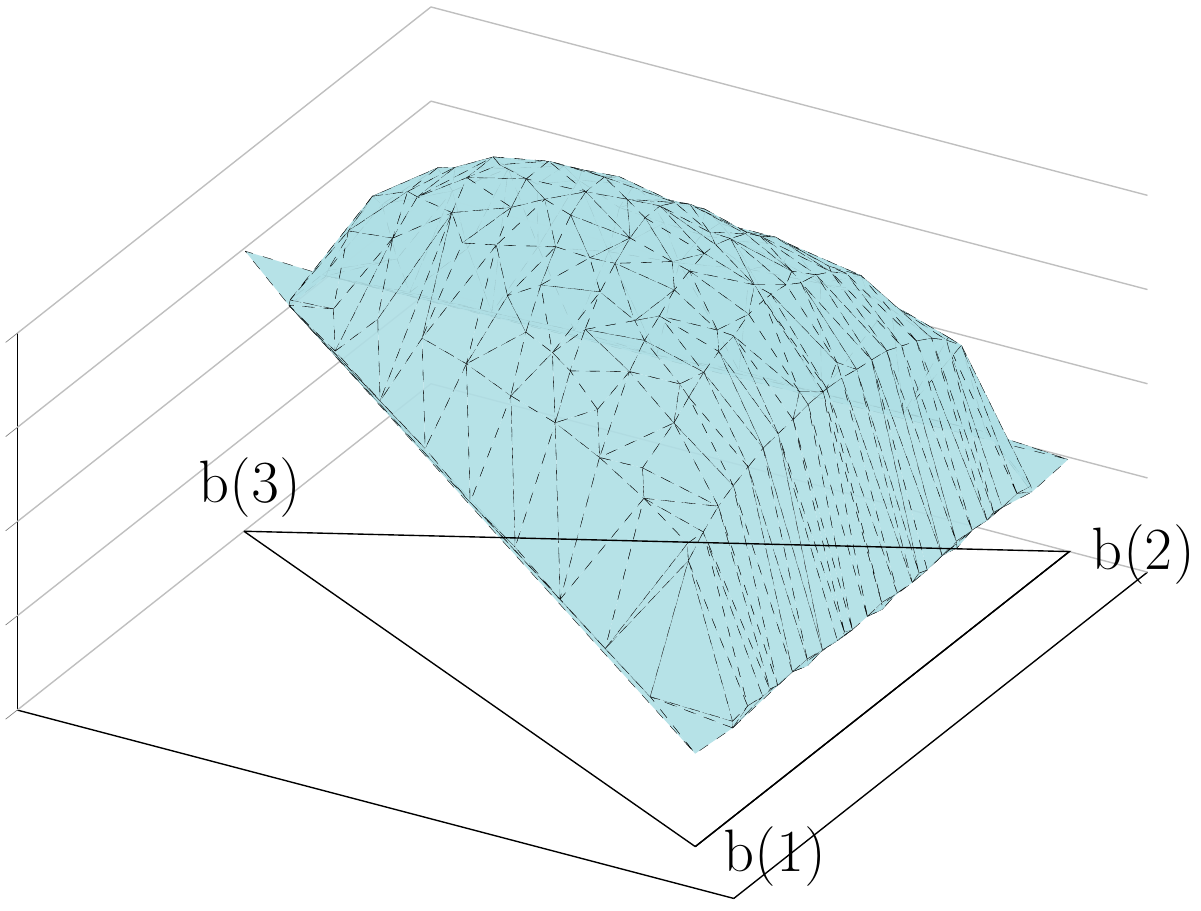}}
        \label{fig:increase_samples_medium}
    \end{subfigure}
    \begin{subfigure}[b]{0.45\textwidth}
        \centering\scalebox{0.45}{
        \includegraphics[]{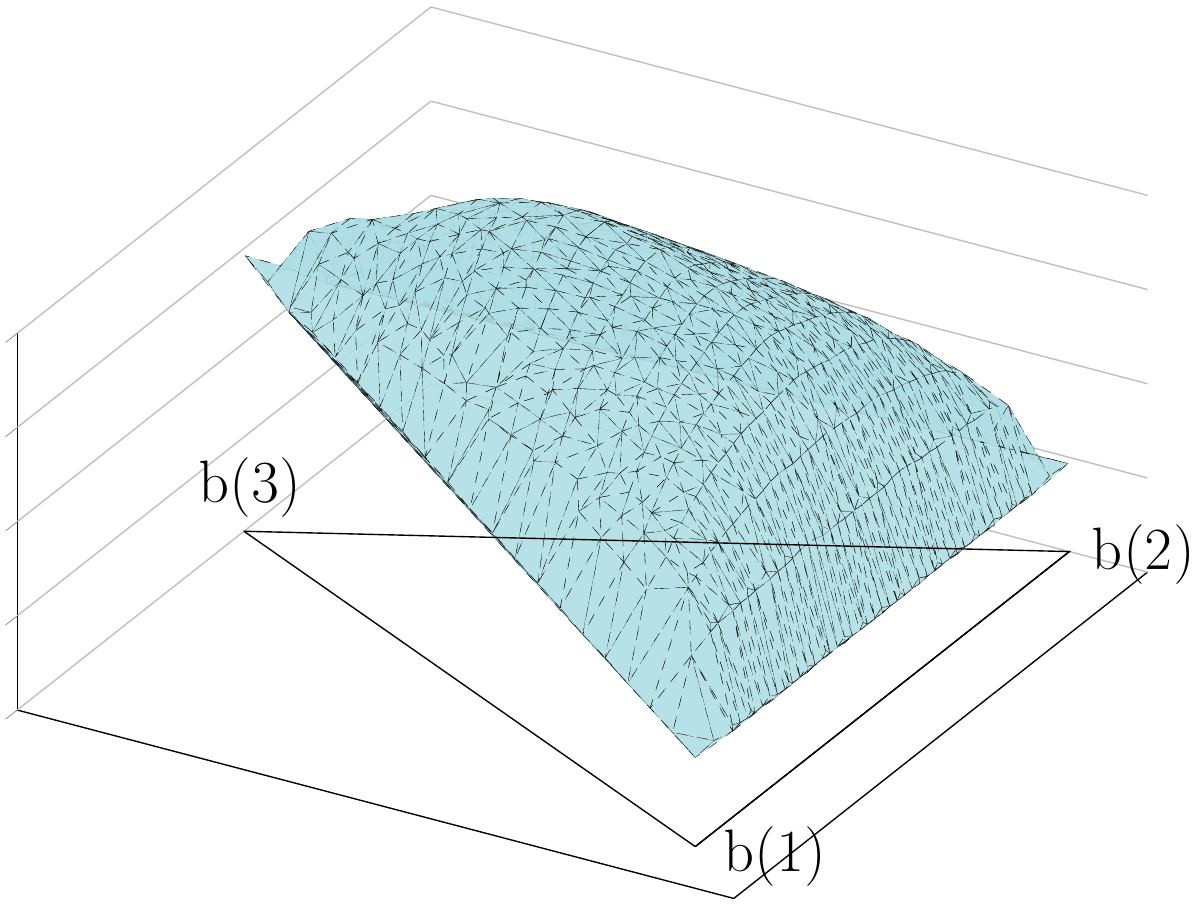}}
        \label{fig:increase_samples_fine}
    \end{subfigure}
    \caption{Prior belief state value functions plotted on the 3D belief simplex  for an increasing number of sampled belief states.}
    \label{fig:increase_samples}
\end{figure}

Following from Lemma \ref{lem:monotonicity}, if it is believed that (\ref{eq:lin_prog-1})-(\ref{eq:modModPosBel}) yield poor approximations, then one can add sample beliefs to $\hat{\mathcal{B}}$, and recompute. Recall, by Assumption 1, that increasing the cardinality of the set $\hat{\mathcal{B}}$ likewise requires the cardinality of the observation alphabet to also increase such that $|\hat{\mathcal{B}}|$$=$$M$ holds. In this sense, Lemma \ref{lem:monotonicity} implies that an agent with an additional sensor can do no worse than an agent without that sensor. Fig. \ref{fig:increase_samples} illustrates that the fixed point $V(b)$ of the value iteration is monotonically non-increasing in $M$ for a simple three-state example (additional information about this example is provided in the Appendix~\ref{appendix:three_state}).

\subsection{Approximation of Value Functions}\label{sec:val_func_convergence}

Theorem~\ref{thm_T} implies that the value function $V^*$ for the main problem \eqref{eq:MainObjective} can be obtained by the value iteration procedure $V_k=T^kV_0$. Unfortunately, such a procedure is computationally intractable. In the previous subsection, we showed that an approximation $\tilde{V}^*$ of $V^*$ can be obtained through the modified value iteration procedure $\tilde{V}_k=\tilde{T}^k\tilde{V}_0$, which can be performed exclusively on the IFBS.
We now study how the gap between $V^*$ and $\tilde{V}^*$ depends on the sample density of the IFBS. First, we characterize the sample density of the IFBS $\hat{\mathcal{B}}\subset \Delta(S)$ as follows:
For each $\hat{b}\in\Delta(S)$, denote by $\pi(\hat{b})\in\hat{\mathcal{B}}$ the nearest element of $\hat{\mathcal{B}}$ from $\hat{b}$ whose support is contained in the support of $\hat{b}$. That is,
\begin{align}
    \pi(\hat{b})\triangleq &\argmin\nolimits_{\hat{b}^m\in\hat{\mathcal{B}}} \quad  \|\hat{b}-\hat{b}^m\|_\infty \label{eq:def_pi_b} \\
    &\quad \text{ s.t. } \quad \text{support}(\hat{b}^m) \subseteq \text{support}(\hat{b}). \nonumber
\end{align}
The constraint that $\text{support}(\hat{b}^m) \subseteq \text{support}(\hat{b})$ will be necessary for the proof of Lemma \ref{lem:noisyvalueiteration} below. Using $\pi(\cdot)$, define the density parameter 
\begin{equation}
\label{eq:def_epsilon_m}
    \hat{\epsilon}\triangleq \max_{\hat{b}\in \Delta(S)}  \|\hat{b} - \pi(\hat{b})\|_{\infty}.
\end{equation}
How well the function $V^*$ can be approximated also depends on the ``regularity" of $V^*$. We define the regularity parameter $\hat{\delta}$ as a positive constant such that
\begin{equation}
\label{eq:def_delta_m}
|V^*(b)-V^*(b')|\leq \hat{\delta}
\end{equation}
holds for all $b, b' \in \Delta(S)$ such that $\|b-b'\|_\infty\leq \hat{\epsilon} |S|$. The main result of this subsection critically relies on the following lemma, which provides an upper bound on the difference between the operators $T$ and $\tilde{T}$ applied to the same function $V$.
For a given function $V\in B(\Delta(S))$, denote by $V|_\mathcal{B} \in \mathbb{R}^M$ the restriction to the set $\mathcal{B}$; i.e., $V|_\mathcal{B}$ is the function $V$ evaluated only at points in the set $\mathcal{B}$.

\begin{lemma}\label{lem:noisyvalueiteration}
Suppose that a function $V\in B(\Delta(S))$ satisfies
\begin{equation}
\label{eq:continuity_hypothesis}
    |V(b)-V(b')|\leq \hat{\delta}
\end{equation}
for all $b, b' \in \Delta(S)$ such that $\|b-b'\|_\infty \leq \hat{\epsilon} |S|$. Then,
\begin{equation}
\label{eq:t_tilde_v_main}
    \|(TV)|_\mathcal{B}-\tilde{T}(V|_\mathcal{B})\|_\infty \leq \epsilon
\end{equation}
where
\begin{align}
    \epsilon&=\gamma \hat{\delta} + \hat{\epsilon} \beta |\log \hat{\epsilon}||S| \nonumber \\
    &\quad +\hat{\epsilon} \Big( \beta\!\!\! \sum_{s\in \bar{S}(b) \neq 0}\!\!\! |\log b(s)|+\sum_{s,a}|C(s,a)| \Big). \label{eq:def_epsilon_thm}
\end{align}
\end{lemma}
\noindent
\textbf{Proof:} See Appendix~\ref{appendix:lem:noisyvalueiteration}. $\Box$

The main result of this subsection is summarized as follows:
\begin{restatable}{thm}{thmconvergence}\label{thm_converge}
Let $\mathcal{B}$ and $\hat{\mathcal{B}}$ be fixed, and define the parameter $\hat{\epsilon}$ by \eqref{eq:def_epsilon_m}.
    Let $V^*\in B(\Delta(S))$ be the unique function satisfying $V^*=TV^*$, and define the parameter $\hat{\delta}$ by \eqref{eq:def_delta_m}. 
    Define the sequence $\tilde{V}_k\in\mathbb{R}^M$, $k=1, 2, ...$ by $\tilde{V}_k =\tilde{T}^k \tilde{V}_0$ where $\tilde{V}_0\in\mathbb{R}^M$ is an arbitrary bounded vector. Then,
    \begin{align}\label{eq:noisy_val_bound2}
        \lim \sup _{k\rightarrow\infty} \|V^*|_\mathcal{B} - \tilde{V}_{k} \|_{\infty} \leq \frac{\epsilon}{1-\gamma}
    \end{align}
    where $\epsilon$ is defined by \eqref{eq:def_epsilon_thm}.
\end{restatable}
\textbf{Proof:} Notice that
\begin{align}
    &\|V^*|_\mathcal{B} - \tilde{V}_{k+1}\|_{\infty} \nonumber \\
    & =\|(TV^*)|_\mathcal{B}-\tilde{T}\tilde{V}_k\|_\infty \nonumber\\
    &= \| (TV^*)|_\mathcal{B}-\tilde{T}(V^*|_\mathcal{B})+\tilde{T}(V^*|_\mathcal{B})-\tilde{T}\tilde{V}_k \|_{\infty} \nonumber \\
    & \leq \|\tilde{T}(V^*|_\mathcal{B})-\tilde{T}\tilde{V}_k \|_{\infty}+ \|(TV^*)|_\mathcal{B}-\tilde{T}(V^*|_\mathcal{B})\|_\infty \nonumber \\
    & \leq \gamma \| V^*|_\mathcal{B} - \tilde{V}_k \|_{\infty} + \epsilon.  \label{eq:nvb2}
\end{align}
The first equality is obtained by invoking $V^*=TV^*$ and $\tilde{V}_{k+1}=\tilde{T}\tilde{V}_k$.
In the last step, we used the fact that $\tilde{T}$ is contractive (Theorem~\ref{thm_T_tilde}) and the result of Lemma~\ref{lem:noisyvalueiteration}.
Define a sequence $e_{k}$ of positive numbers by
\begin{align}\label{eq:pos_num_seq2}
    e_{k+1} = \gamma e_{k} + \epsilon,
\end{align}
taking $e_{0}=\| V^*|_\mathcal{B} - \tilde{V}_{0} \|_{\infty}$. Then,
\begin{align}\label{eq:pos_seq_infty2}
    \lim_{k\rightarrow\infty} e_{k} = \frac{\epsilon}{1 - \gamma}.
\end{align}
Combining the results of \eqref{eq:nvb2}, \eqref{eq:pos_num_seq2}, and \eqref{eq:pos_seq_infty2}, it is straightforward to show by induction that
\begin{align*}
    \| V^*|_\mathcal{B} - \tilde{V}_k \|_{\infty} \leq e_{k} \quad \forall k = 0,1,\ldots,
\end{align*}
from which \eqref{eq:noisy_val_bound2} follows.
$\Box$

Notice that the constant $\epsilon$ appearing in \eqref{eq:noisy_val_bound2} depends on $\hat{\epsilon}$ and $\hat{\delta}$.
Since the optimal value function $V^*$ is not known in advance, it is in general not possible to compute $\hat{\delta}$.
However, in circumstances where $V^*$ is known to be uniformly continuous, for each $\hat{\delta}>0$, the condition \eqref{eq:def_delta_m} can always be guaranteed by choosing a sufficiently small $\hat{\epsilon}>0$, i.e., by making the set $\hat{\mathcal{B}}$ sufficiently dense in $\Delta(S)$.
Therefore, in such cases, the vector $\tilde{V}^*$ can approximate $V^*|_\mathcal{B}$ arbitrarily well by increasing the sample density of $\hat{\mathcal{B}}$ in $\Delta(S)$. 

It is currently not known under what conditions the uniform continuity of $V^*$ is guaranteed. Obtaining these conditions remains the subject of future work.

\section{Numerical Example: Mars Rover}

\begin{figure}[t]
    \centering\scalebox{0.55}{
    \includegraphics[]{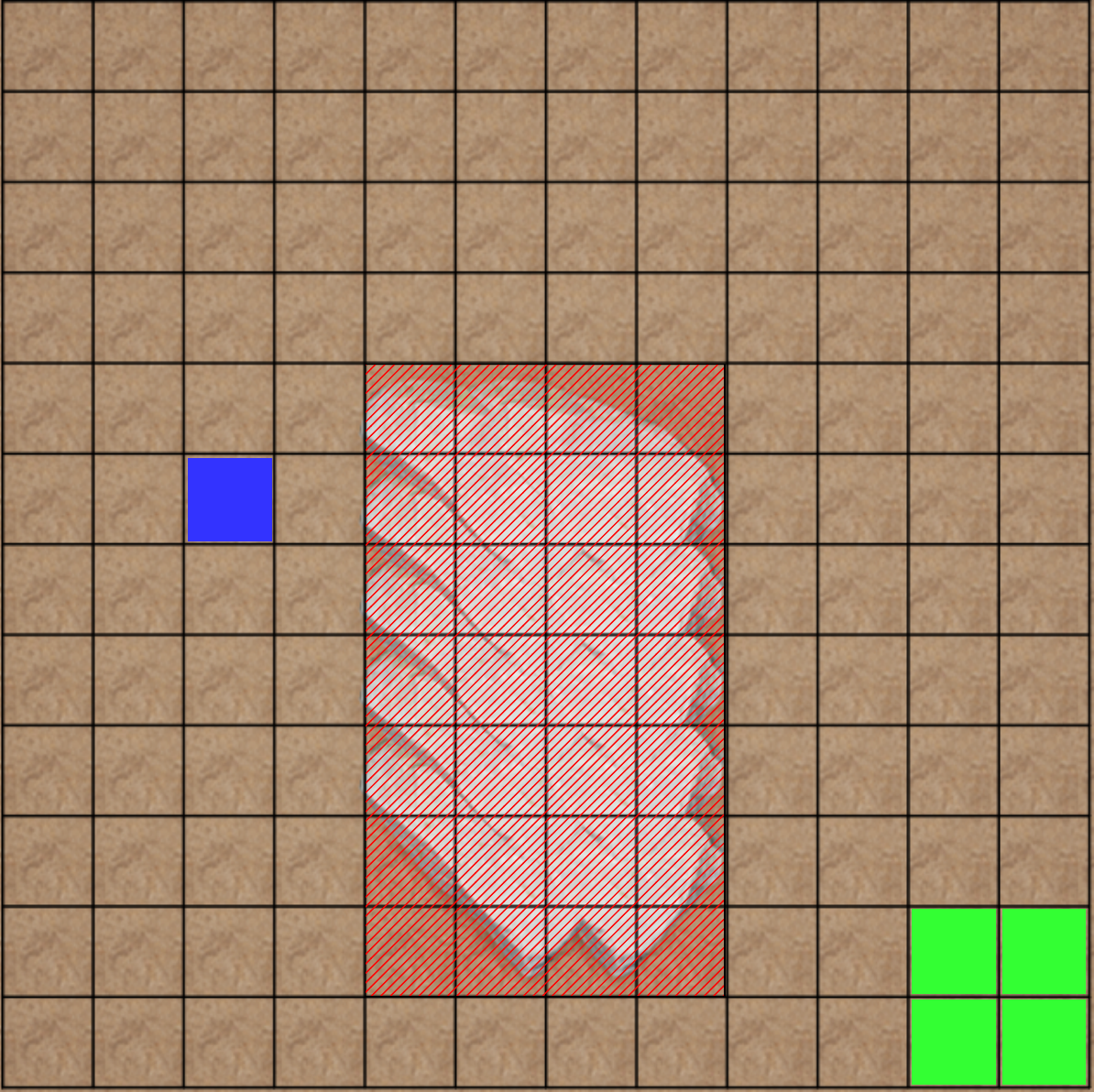}}
    \caption{Mars gridworld environment considered.}
    \label{fig:mars_gridworld}
\end{figure}

To demonstrate the SPADE framework, we consider a Mars rover that must complete a surveying task by maneuvering from its initial position to a target position. To reach the target position, the rover must avoid a dangerous, rocky section in the center of its environment. To accomplish this objective, the rover can choose from one of several possible paths: it can either follow a more direct, but dangerous, path by travelling towards the bottom of the environment, or it can follow a longer, but safer, path around the top of the rocky area. In this example, we study the relation between the relative cost of perception and the resulting path that the rover follows. In this context, the notion of ``task-relevant" information pertains to observations identifying the rover's underlying state in the environment. The environment of the rover is shown in Fig. \ref{fig:mars_gridworld}, where the blue state is the rover's initial position, the green states represent the target area that the rover must travel to, and the red dashed states are the rocky states to be avoided.

In each state in the environment, the rover can choose between one of 4 available actions: move one step either to the left, to the right, up, or down. Due to stochasticity, however, the rover may either remain in its current state or slip into one of the other surrounding states, each with probability $\nicefrac{0.05}{8}$. If the rover were to transition to a state outside of its environment, it instead transitions to the closest state still within its environment. In Fig. \ref{fig:mars_gridworld}, the green and red-dashed states are absorbing; i.e., if the rover reaches any of its target states or one of the rocky states, it remains there. Furthermore, the green states have no associated environmental cost for taking any action within them. For all other states, the rover incurs a cost of $1$ for taking any action.

We sample six posterior belief states for each state $s$ in the environment. For the first posterior belief state, we set $\hat{b}(s)$$=$$1$ and $0$ otherwise for each $s$$\in$$S$, satisfying Assumption \ref{assumption1_EP}. For the second posterior belief state, we set $\hat{b}(s)$$=$$0.5$ and $\hat{b}(s')$$=$$\nicefrac{0.5}{8}$ for all $s'$, $s'$$\neq$$s$ in the $3$$\times$$3$ square centered around state $s$. For the third posterior belief state, we repeat the previous using $\hat{b}(s)$$=$$0.75$. We follow a similar process now considering the $5$$\times$$5$ square centered around state $s$. First, we set $\hat{b}(s)$$=$$0.5$, $\hat{b}(s')$$=$$\nicefrac{0.5}{16}$ for the eight states $s'$ in the $3$$\times$$3$ square around state $s$, and $\hat{b}(s'')$$=$$\nicefrac{0.5}{32}$ for the remaining sixteen states in the $5$$\times$$5$ square. We then repeat this process using $\hat{b}(s)$$=$$0.35$ and $\hat{b}(s)$$=$$0.20$. If this procedure would allocate non-zero probability mass to a state outside the environment, it is instead allocated to the closest state within the environment. This procedure yields a set of $864$ posterior belief states, which subsequently yields a set of $3456$ prior belief states obtained by using (\ref{eq:predictEquation}).

We use a discount factor of $\gamma$$=$$0.95$ and consider two values of the weighting factor $\beta$: $\beta$$=$$0$ and $\beta$$=$$20$. A value of $\beta$$=$$0$ corresponds to a situation where the rover incurs no perception cost. Considering an infinite time horizon, we perform value iteration until convergence for each value of $\beta$, wherein we solve (\ref{eq:lin_prog-1})-(\ref{eq:lin_prog-3}) for each prior belief state and (\ref{eq:modModPosBel}) for each posterior belief state at each iteration. To solve each LP, we use the default LP solver available in MATLAB \cite{MatlabOTB}.

\begin{figure*}
    \centering
    \begin{subfigure}[b]{\textwidth}
        \includegraphics[width=\textwidth]{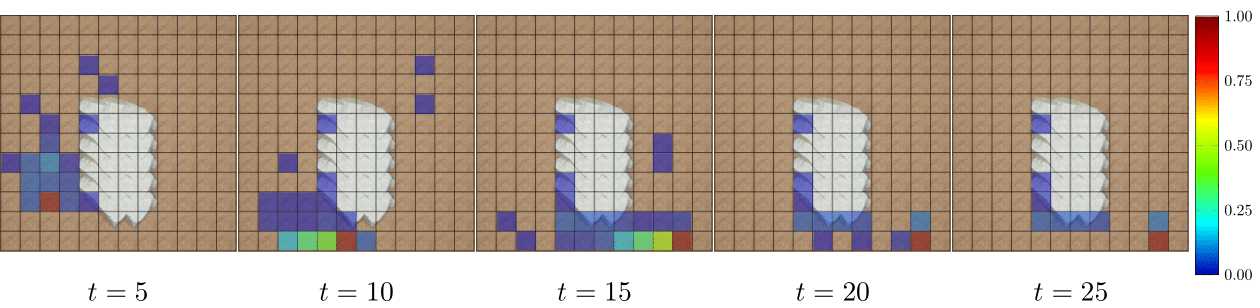}
        \caption{Case 1: $\beta=0$. The agent incurs no perception costs and is free to synthesize perception strategies that yield perfect state information. The agent is able to leverage this information in order to take a more direct path to the goal state.}
        \label{fig:gamma_0}
    \end{subfigure}
    \begin{subfigure}[b]{\textwidth}
        \centering
        \includegraphics[width=\textwidth]{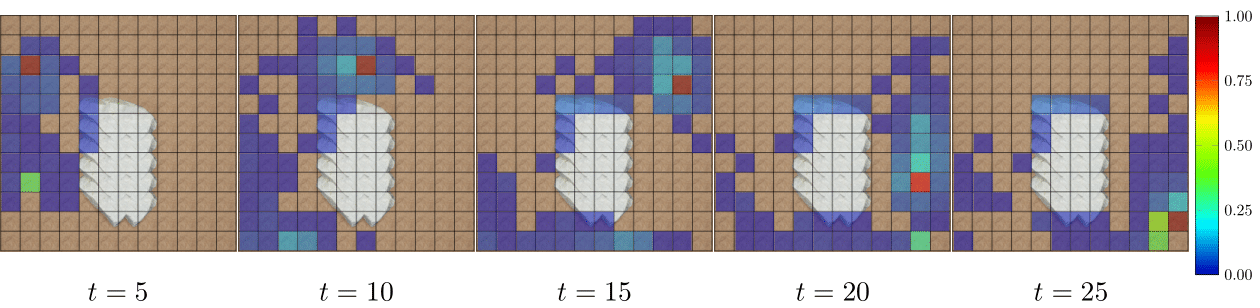}
        \caption{Case 2: $\beta=20$. The agent faces prohibitive perception-related costs relative to Case 1. For this reason, the agent instead favors a longer path around the rocky obstacle. This path does not require high-fidelity observations, reducing perception costs.}
        \label{fig:gamma_20}
    \end{subfigure}
    \caption{Expected rover state residences averaged over $1000$ trials for $\beta = 0$ and $\beta = 20$ plotted at various time steps. The shading of each state corresponds to the fraction of the trials that occupied that state at the given time step.}
    \label{fig:gamma_time_horizon}
\end{figure*}

To discuss the qualitative differences between the synthesized joint perception-action strategies for the values of $\beta$ considered, we examine the sample trajectories that they generate. Fig. \ref{fig:gamma_time_horizon} plots the expected state residence averaged over $1000$ trials for each value of $\beta$ considered. In the case that $\beta$$=$$0$, we see from Fig. \ref{fig:gamma_0} that the rover takes the shorter path underneath the rocky obstacles to reach its objective, as, in the absence of perception costs, it is strictly focused on minimizing its remaining environmental costs. The rover is able to follow this path as it has perfect knowledge of its state in the environment. Because it incurs no perception costs, the synthesized perception strategy will always uniquely indicate the true underlying state of the rover. The rover is thus able to leverage this perfect state information towards taking a more direct path to reach the target states.

In the case that $\beta$$=$$20$, perception costs have driven the rover to exhibit different behavior. In the majority of the simulations, the rover takes the longer path around the top of the rocky obstacles, typically providing itself at least a row of separation around them. Intuitively, the synthesized perception strategy drives the rover to maintain a diffuse belief state, as such belief states correspond to lower perception costs. Thus, to balance perception and environmental costs, the rover remains in belief states that are diffuse yet have a low, if not zero, probability of residing in a rocky state. In some cases; however, the rover initially slips several states in the opposite direction of the safer path. Once in such a state, the environmental costs associated with following an even longer (but safer) trajectory begin to dominate the perception costs. In these cases, the rover follows trajectories more similar to those displayed in the case of $\beta$$=$$0$.

Through this example, we see how the relative costs associated with perception and the environment of the agent can lead to significantly different behavior.

\section{Conclusion and Future Work}

We considered a simultaneous perception-action design problem for an agent wherein the perception costs were modelled using the directed information. The agent's objective function was decomposed into two coupled sets of recursive Bellman equations, which allowed us to obtain a tractable, approximate solution through a novel \textit{method of invariant finite belief sets}. The proposed method restricts the agent to operate exclusively on a finite subset of the continuous belief space. An optimal simultaneous perception-action strategy can then be obtained using a backward dynamic programming approach wherein a linear program is solved for each prior belief state at each iteration. Future work must consider the validity of the assumption on the structure of the continuous state-space value functions used to derive the convergence result.

Several natural extensions of the SPADE framework are as follows. To start with, once the optimal perception strategy $\mathcal{P}(z|s,b)$ has been obtained, the next step is to select, or develop, a sensor that ``physically realizes" the perception strategy (at least approximately). The types of additional constraints that must be imposed on the perception strategy to allow for such a sensor remains the subject of future work. For example, one must consider cases where no sensing device is available to distinguish some states from one another, or the case that certain observations are only available in specific states.

Furthermore, for the tabular algorithm we propose in equations (12)-(13), naïvely constructing the posterior belief set $\hat{\mathcal{B}}$ yields an impractical cardinality for realistic, large-scale problems. Developing methods to cleverly construct $\hat{\mathcal{B}}$ remains an important research opportunity for future study. An alternative direction for mitigating computational costs is to work on the feature space rather than the original state space. Doing so may additionally faciliate the incorporation of temporal logic into the SPADE framework, allowing for the expression of more complicated tasks and objectives.

\bibliography{ref.bib}

\appendix

\section{Details on Information-theoretic Perception Cost}
\label{sec:appendix_perception_cost}

We start by providing a more rigorous rationale behind our choice of the directed information, $I(\bs_{1:T}$$\rightarrow$$\bz_{1:T})$, for our perception cost. To this end, notice that
\begin{subequations}
\begin{align}
&I(\bs_{1:T}\rightarrow \bz_{1:T}) =\sum\nolimits_{t=1}^T I(\bs_{1:t};z_{t}|\bz_{1:t-1}) \tag{23a} \\
&=\sum\nolimits_{t=1}^T I(\bs_t;\bz_t|\bz_{1:t-1})-I(\bs_{1:t-1};\bz_t|\bs_t,\bz_{1:t-1}) \label{eq:di2} \tag{23b}\\
&=\sum\nolimits_{t=1}^T I(\bs_t;\bz_t|\bz_{1:t-1}) \tag{23c}\\
&=\sum\nolimits_{t=1}^T H(\bs_t|\bz_{1:t-1})-H(\bs_t|\bz_{1:t})\label{eq:di4}. \tag{23d}
\end{align}
\end{subequations}
The second term in \eqref{eq:di2} is zero as our model assumes that $\bz_t$ is independent of $\bs_{1:t-1}$ given $\bs_t$, as shown in Fig.~\ref{fig:Perception_diagram}. Therefore, the directed information is equivalent to the summation of the stage-additive information gains; i.e., the difference in entropy of the state variable before and after incorporating the newest measurement $\bz_t$.
We interpret this information gain as the minimum number of information \emph{bits} that must be delivered from the perception unit to the action unit in each time step. The set of SPADE parameters that minimize $I(\bs_{1:T}$$\rightarrow$$\bz_{1:T})$ are advantageous, since an optimal source coding (i.e., data compression) scheme can potentially reduce the data traffic from the perception unit to the action unit to $I(\bs_{1:T}\rightarrow \bz_{1:T})$ bits. 

We now consider a formal analysis that allows us to provide the directed information a Shannon-theoretic operational meaning. To this end, we introduce a model in which the communication channel from the perception unit to the action unit is a noiseless bitpipe, through which the message $\bz_t$ is delivered in the form of a variable-length, uniquely decodable binary code.

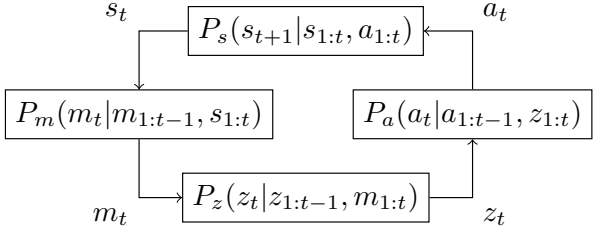
\begin{figure}[t]
  \centering
  \scalebox{0.9}{
    \begin{tikzpicture}

\node (Ps) at (0,0) [rectangle,draw] {$P_{s}(s_{t+1}|s_{1:t},a_{1:t})$};
\node (Pm) at (-2,-1) [rectangle,draw] {$P_{m}(m_{t}|m_{1:t-1},s_{1:t})$};
\node (Pz) at (0,-2) [rectangle,draw] {$P_{z}(z_{t}|z_{1:t-1},m_{1:t})$};
\node (Pa) at (2,-1) [rectangle,draw] {$P_{a}(a_{t}|a_{1:t-1},z_{1:t})$};

\draw[->] (Ps) -| node[above left] {$s_{t}$} (Pm);
\draw[->] (Pm) |- node[below left] {$m_{t}$} (Pz);
\draw[->] (Pz) -| node[below right] {$z_{t}$} (Pa);
\draw[->] (Pa) |- node[above right] {$a_{t}$} (Ps);

\end{tikzpicture}}
    \caption{Feedback system considered.}
  \label{fig:DPI}
\end{figure}

Let $P_s, P_m, P_z$ and $P_a$ be conditional probability distributions with the structures shown in Fig.~\ref{fig:DPI}. Given an initial distribution $P_s(\bs_1)$, let $\bs_{1:T}, \bm_{1:T}, \bz_{1:T}$ and $\ba_{1:T}$ be random processes defined by the feedback diagram shown in Fig.~\ref{fig:DPI}. Then, we can obtain the following lemma.

\begin{figure}[t]
    \centering
    \begin{tikzpicture}

\node (T) at (0,0) [rectangle,draw] {$T(s_{t+1}|s_{t},a_{t})$};
\node (P) at (-3,-1.75) [rectangle,draw] {$P(m_{t}|s_{t})$};
\node (Zba) at (3,-1.75) [rectangle,draw] {{\begin{tabular}{ll}
     & $z_{t}=d(m_{t})$ \\
     & $\hat{b}_{t}=f(\hat{b}_{t-1},a_{t-1},z_{t})$ \\
     & $a_{t}=g(\hat{b}_{t})$
\end{tabular}}};

\draw[->] (T) -| node[above left] {$s_{t}$} (P);
\draw[->] (P) -- node[above] {$m_{t}\in\{0,1\}^{l_{t}}$} (Zba);
\draw[->] (Zba) |- node[above right] {$a_{t}$} (T);

\end{tikzpicture}
        \caption{Perception through bitpipe.}
    \label{fig:bitpipe}
\end{figure}
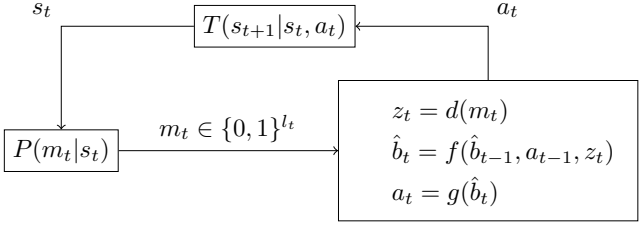

\begin{lemma}\label{Lemma_pdi}
(Data processing inequality for directed information) Given the Feedback system shown in Fig. \ref{fig:DPI}, the following inequalities hold:
\begin{align*}
&I(\bs_{1:T}\rightarrow \ba_{1:T}) \leq I(\bs_{1:T}\rightarrow \bz_{1:T}) \leq I(\bm_{1:T}\rightarrow \bz_{1:T}) \\
&I(\bs_{1:T}\rightarrow \ba_{1:T}) \leq I(\bm_{1:T}\rightarrow \ba_{1:T}) \leq I(\bm_{1:T}\rightarrow \bz_{1:T}).
\end{align*}
\end{lemma}
\noindent
\textbf{Proof:}
See, e.g., \cite{derpich2021directed}.
$\Box$

To provide the directed information a Shannon-theoretic operational meaning, assume that messages from the perception unit to the action unit are communicated through a noiseless bitpipe, as shown in Fig.~\ref{fig:bitpipe}, and that delivering an individual bit incurs a unit cost. This formulation provides a meaningful model to estimate the cost of communication in applications where sensor data is transmitted over a digital communication channel.
In time step $t$, the perception unit produces a uniquely decodable variable-length binary code $m_t$$\in$$\{0,1\}^{\ell_t}$, where $\ell_t$ is the length of the code.
The message $m_t$ is decoded in the action unit to reproduce the observation signal $z_t$$\in$$Z$.
Since communication is costly, the design goal of the simultaneous perception-action system, including message encoder and decoder, is to minimize 
\begin{equation}\tag{24}
    \sum\nolimits_{t=1}^{T} \mathbb{E}C(\bs_t,\ba_t)+\mathbb{E}[\bl_t],
\end{equation}
where $\mathbb{E}[\bl_t]$ is the expected codeword length.
Although it is difficult to evaluate $\mathbb{E}[\bl_t]$ directly, it can be approximated using directed information, as shown in the following lemma.

\begin{lemma}\label{Lemma_code}
For the communication system shown in Fig.~\ref{fig:bitpipe}, we have
\begin{equation}\tag{25}
I(\bs_{1:T}\rightarrow \bz_{1:T})\leq \sum\nolimits_{t=1}^T\mathbb{E}[\bl_t]. \label{eq:di_lb}
\end{equation}
\end{lemma}
\noindent
\textbf{Proof:} The following chain of inequalities establishes the claim:
\begin{subequations}
\begin{align}
    &I(\bs_{1:t}\rightarrow \bz_{1:t})\leq I(\bm_{1:t}\rightarrow \bz_{1:t}) \label{eq:lem_code1} \tag{26a} \\
    &=\sum\nolimits_{t=1}^T I(\bm_{1:t};\bz_t|\bz_{1:t-1}) \nonumber \\
    &=\sum\nolimits_{t=1}^T I(\bm_t;\bz_t|\bz_{1:t-1})- \underbrace{I(\bm_{t-1};\bz_t|\bm_t, \bz_{1:t-1})}_{=0} \nonumber \\
    &=\sum\nolimits_{t=1}^T H(\bm_t|\bz_{1:t-1})-\underbrace{H(\bm_t|\bz_{1:t})}_{\geq 0} \nonumber \\
    &\leq \sum\nolimits_{t=1}^T H(\bm_t) \label{eq:lem_code2} \tag{26b} \\
    &\leq \sum\nolimits_{t=1}^T \mathbb{E}[\bl_t]. \label{eq:lem_code3} \tag{26c}
\end{align}
\end{subequations}
The data processing inequality (Lemma~\ref{Lemma_pdi}) is applied to \eqref{eq:lem_code1}. The inequality \eqref{eq:lem_code2} holds due to the fact that conditioning can only reduce the entropy. The final step \eqref{eq:lem_code3} follows from the fact that any uniquely decodable code is a uniquely decodable code of itself, and thus its expected codeword length is lower-bounded by its entropy, see, e.g., (Theorem 5.3.1) of \cite{cover2012elements}.
$\Box$

Evaluating the tightness of the lower bound \eqref{eq:di_lb} is more challenging. However, it is reported \cite{tanaka2017lqg, kostina2019rate} that the construction of source coders operating at a rate close to this lower bound is possible under some special circumstances, such as in Linear-Quadratic-Gaussian settings.

\section{Proofs of Lemmas \ref{Lemma1}-\ref{lem:monotonicity}}\label{sec:theory_proofs}

\feassolution*
\textbf{Proof:} Starting with the constraint given in (8), we sum over all $m$$=$$1\ldots M$ to obtain 
\begin{equation}\label{eq:Lemma1-Eq1}\tag{14}
    \sum\nolimits_{m=1}^{M} \alpha^{m}_{b} \hat{b}^{m} = \sum\nolimits_{m=1}^{M} \text{diag}(b) \mathcal{P}^{m}_{b}.
\end{equation}
Because $\text{diag}(b)$ is present within each term, we can move it outside the summation. Then, noting that the perception strategy for each  state $s$ must be a valid probability distribution; i.e., $\sum_{m=1}^{M}$$\mathcal{P}(z^{m}|s,b)$$=$$1$ for each $s$$\in$$S$, we must have that $\sum_{m=1}^{M}\mathcal{P}_{b}^{m}$$=$$\mathbf{1}_{|S|\times 1}$. Substituting this condition into (\ref{eq:Lemma1-Eq1}) yields
\begin{equation}\label{eq:Lemma1-Eq2}\tag{15}
    \sum_{m=1}^{M} \alpha^{m}_{b} \hat{b}^{m} = \text{diag}(b) \mathbf{1}_{|S|\times 1},
\end{equation}
which we can write more intuitively in matrix notation as
\begin{equation}\label{Lemma1Eq3}\tag{16}
    \begin{bmatrix}
        | & { } & | \\
        \hat{b}^{1} & \cdots & \hat{b}^{M} \\
        | & { } & |
    \end{bmatrix}
    \begin{bmatrix}
        \alpha_{b}^{1} \\
        \vdots \\
        \alpha_{b}^{M}
    \end{bmatrix} = 
    \begin{bmatrix}
        | \\
        b \\
        |
    \end{bmatrix}.
\end{equation}
Because we chose $\hat{\mathcal{B}}$ such that Assumption 1 is satisfied, we can express any $b$ as a convex combination of extreme points of $\Delta(S)$. Thus, there exist non-negative coefficients $\alpha_{b}^{m}$, $m$$=$$1\ldots M$ satisfying (\ref{Lemma1Eq3}). Furthermore, in realistic applications, we will often have that $M$$\gg$$|S|$. By this condition, there exist infinitely many solutions to (9)-(10). $\Box$

\linearprogram*

\textbf{Proof:}
Considering only the states
$s\in \overline{S}(b)$, we can use~\eqref{eq:vectorForm} to parameterize the perception strategy variables as
%
\begin{equation}\label{eq:sm_lemma_2_param}\tag{17}
    \mathcal{P}(m|s,b) = \alpha_{b}^{m}
    \frac{\hat{b}^{m}(s)}{b_{t}(s)},
\end{equation}
for all $s$$\in$$\overline{S}(b)$ and for all $m$$=$$1,\ldots,M$. Since $b$ and $\hat{b}^{m}$, $m$$=$$1,\ldots,M$, are nonnegative vectors, and $\alpha_{b}^{m}$ is constrained to be nonnegative, the parameterized perception strategy variables are likewise guaranteed to be nonnegative. Substituting this parameterization for the perception strategy variables into the first set of linear constraints in~\eqref{eq:linearConstraints} and multiplying each side by $\text{diag}(b[\overline{S}(b)])$, we obtain
%
\begin{align}\label{eq:sm_lemma_2_cons_sub}\tag{18}
    \sum\nolimits_{m\in \overline{M}(b)} \alpha_{b}^{m} \hat{b}^{m}[\overline{S}(b)] = b[\overline{S}(b)]
\end{align}
Now, substituting the parameterization for the perception strategy variables into~\eqref{eq:modifiedBellman}, we see that
\begin{subequations}
    \begin{align}
        & \sum_{m=1}^{M} \sum_{s\in \overline{S}(b)} b(s) \mathcal{P}(m|s,b) \cdot \ldots\nonumber \\
        & \qquad \qquad \qquad \left( \beta \log\frac{\mathcal{P}(m|s,b)}{\alpha_{b}^{m}} + \hat{V}(\hat{b}^{m}) \right) \tag{19a} \\
        & = \sum_{m=1}^{M}\sum_{s\in \overline{S}(b)} \alpha_{b}^{m}\hat{b}^{m}(s) \left( \beta \log\frac{\hat{b}^{m}(s)}{b(s)} + \hat{V}(\hat{b}^{m}) \right) \tag{19b} \\
        & = \sum_{m=1}^{M} \alpha_{b}^{m} \sum_{s\in \overline{S}(b)} \hat{b}^{m}(s) \left(\beta\log\frac{\hat{b}^{m}(s)}{b(s)} + \hat{V}(\hat{b}^{m}) \right) \tag{19c} \\
        & = \sum_{m=1}^{M} \alpha_{b}^{m} \left(\beta \sum_{s\in \overline{S}(b)} \hat{b}^{m}(s)\log\frac{\hat{b}^{m}(s)}{b(s)} + \hat{V}(\hat{b}^{m}) \right). \label{eq:sm_lemma_2_obj} \tag{19d}
    \end{align}
\end{subequations}
By defining
%
\begin{align*}
        & F_{m} = \beta D(\hat{b}^{m}||b) + \hat{V}(\hat{b}^{m}), \\
        & D(\hat{b}^{m}||b) = \sum\nolimits_{s\in \overline{S}(b)}\hat{b}^{m}(s) \log \frac{\hat{b}^{m}(s)}{b(s)}.
    \end{align*}
we can rewrite the objective function in (\ref{eq:sm_lemma_2_obj}) in tandem with the constraints in (\ref{eq:sm_lemma_2_cons_sub}) to obtain the desired LP,
%
%
completing the proof. $\Box$

\monotonicity*
\textbf{Proof:} Consider a perception strategy $\mathcal{P}$$\in$$\mathcal{P}(b$$\rightarrow$$ \hat{\mathcal{B}})$ with individual observation probabilities $\mathcal{P}(z^{m}|s,b)$, where the observation alphabet has cardinality $M$; i.e., $|Z|$$=$$M$. Now, consider a perception strategy $\mathcal{P}'$ in which the cardinality of the observation alphabet is increased to $M$$+$$1$. Let us construct the new perception strategy $\mathcal{P}'$ in the following manner. For each $z^{m}$, $m$$=$$1,\ldots,M$, let $\mathcal{P}'(z^{m}|s,b)$$=$$\mathcal{P}(z^{m}|s,b)$ for all $S$$\times$$\mathcal{B}$, and let all remaining $\mathcal{P}'(z^{m}|s,b)$ be arbitrarily chosen such that $\mathcal{P}'$ remains in the set $\mathcal{P}'(b$$\rightarrow$$\hat{\mathcal{B}}')$. Then, there is a one-to-one correspondence between the expectations over successor states for both the prior and posterior belief states in the sets $\mathcal{B}$ and $\hat{\mathcal{B}}$, meaning that their respective value functions are equal. Since we chose $\mathcal{P}$ arbitrarily, the value functions for each belief state when synthesizing a perception strategy with an observation alphabet containing $M$$+$$1$ elements cannot be greater than that of the case of synthesizing a perception strategy that contains $M$ elements; i.e.,
\begin{align*}
     & V(b) \geq V'(b), \forall b\in \mathcal{B} \\
     & V(\hat{b}) \geq V'(\hat{b}), \forall b\in \hat{\mathcal{B}},
\end{align*}
completing the proof. $\Box$

\section{Proof of Lemma~\ref{lem:noisyvalueiteration}}
\label{appendix:lem:noisyvalueiteration}

We will utilize the following basic lemma:
\begin{restatable}{lemma}{boundentropy}\label{lem:upper_bound_entropy2}
    Let $p,q\in \Delta(S)$ be two probability distributions such that $\|p-q\|_{\infty} \leq \epsilon \leq \frac{1}{2}$. Then, it holds that
    \begin{align*}
        |H(p) - H(q)| \leq \epsilon |\log \epsilon | |S|.
    \end{align*}
\end{restatable}
\textbf{Proof:} The proof follows that of Theorem 17.3.3 of \cite{cover2012elements}. Consider the concave function $f(t) = -t\log t$. Since $f(0) = f(1) = 0$, it follows that $f(t) \geq 0$ for all $t\in [0,1]$.

The maximum absolute slope of the chord of the function $f(t)$ from $t$ to $t+\epsilon$ is obtained at either end, where either $t=0$ or $t=1-\epsilon$. Thus, for $0\leq t \leq 1-\epsilon$, it follows that
\begin{align}
    |f(t) - f(t+\epsilon)| & \leq \max \{ f(\epsilon), f(1-\epsilon) \} \nonumber \\
    & = -\epsilon \log \epsilon \label{eq:upper_bound_entropy},
\end{align}
since $\epsilon \leq \frac{1}{2}$. Then,
\begin{subequations}
    \begin{align}
        |H(p) & - H(q)| = |\sum_{x\in X} (-p(x) \log p(x) \nonumber \\ 
        & \qquad \qquad \quad \qquad \qquad + q(x) \log q(x) ) | \\
        & \leq \sum_{s \in S} |-p(x) \log p(x) + q(x) \log q(x) \\
        & \leq \sum_{s \in S} -\epsilon \log \epsilon = -\epsilon \log \epsilon |S|\label{eq:upper_bound_entr_final},
    \end{align}
\end{subequations}
where \eqref{eq:upper_bound_entr_final} follows from \eqref{eq:upper_bound_entropy}. $\Box$

To show \eqref{eq:t_tilde_v_main}, we need to prove that the gap between
\begin{align}
    (TV)(b)
    &=\inf_{\mathcal{A},\mathcal{P}}\{\beta R(b,\mathcal{P})+\mathbb{E}_b^{\mathcal{A},\mathcal{P}}[C(\bs,\ba)]\nonumber \\
    &\qquad \qquad+\gamma \mathbb{E}_b^{\mathcal{A},\mathcal{P}} [V(b^{\bz,\ba})] \}  \label{eq:vi_gap_lhs}
\end{align}
and 
\begin{align}
    (\tilde{T}V|_\mathcal{B})(b) &=\min_{\mathcal{A},\mathcal{P}\in \mathcal{P}(b\rightarrow \hat{\mathcal{B}})}\{\beta R(b,\mathcal{P})+\mathbb{E}_b^{\mathcal{A},\mathcal{P}}[C(\bs,\ba)] \nonumber \\
    &\qquad \qquad +\gamma \mathbb{E}_b^{\mathcal{A},\mathcal{P}} [V(b^{\bm,\ba})] \} \label{eq:vi_gap_rhs}
\end{align}
is bounded by $\epsilon$ for each $b\in\mathcal{B}$.
Let $(\mathcal{A}^*,\mathcal{P}^*)$ be a minimizer for \eqref{eq:vi_gap_lhs}. If a minimizer does not exist, one can instead consider an $\epsilon_0$-suboptimal solution for a sufficiently small $\epsilon_0>0$. In this case, the following proof can be adapted with only minor adjustments. Since the perception policy $\mathcal{P}^*(z|s,b)$ is unconstrained, the posterior belief $\hat{b}^z$ can take general values in $\Delta(S)$. In particular, $\mathcal{P}^*(z|s,b)$ drives the prior belief $b$ to a posterior belief $\hat{b}^z\in \Delta(S)$ with probability
\[
\alpha_b^z=\sum_{s\in S} \mathcal{P}^*(z|s,b)b(s).
\]
For each $s\in S$ such that $b(s)\neq 0$, it follows from Bayes' rule \eqref{eq:probUpdate} that $\mathcal{P}^*(z|s,b)$ can be expressed as
\begin{equation}
\label{eq:p_star_alpha_expression}
\mathcal{P}^*(z|s,b)=\alpha_b^z\frac{\hat{b}^z(s)}{b(s)}.
\end{equation}
Notice that $(\mathcal{A}^*,\mathcal{P}^*)$ may not be an admissible policy for \eqref{eq:vi_gap_rhs}.
Instead, we construct an admissible policy $(\tilde{\mathcal{A}}^*,\tilde{\mathcal{P}}^*)$ for \eqref{eq:vi_gap_rhs} (i.e., $\tilde{\mathcal{P}}^*\in \mathcal{P}(b\rightarrow \hat{\mathcal{B}})$) such that $(\tilde{\mathcal{A}}^*,\tilde{\mathcal{P}}^*)$ is ``close" to $(\mathcal{A}^*,\mathcal{P}^*)$.
We construct this perception policy $\tilde{\mathcal{P}}^*\in \mathcal{P}(b\rightarrow \hat{\mathcal{B}})$ from $\mathcal{P}^*$ as follows: 
for each $\hat{b}^m\in\mathcal{\hat{B}}$, define the neighborhood in $Z$ as
\[
N(\hat{b}^m)=\{ z\in Z: \pi(\hat{b}^z)=\hat{b}^m\}.
\]
Clearly, we have $Z=\bigcup_{m=1}^M N(\hat{b}^m)$. Set
\begin{equation}
    \label{eq:def_alpha_m}
    \alpha_b^m=\int_{N(\hat{b}^m)} \alpha_b^z dz,
\end{equation}
for each $m=1,2, ... , M$. Now, define $\tilde{\mathcal{P}}^*$ as
\begin{equation}
    \label{eq:def_p_tilde}
    \tilde{\mathcal{P}}^*(m|s,b)=\begin{cases}
    \alpha_b^m\frac{\hat{b}^m(s)}{b(s)} & \text{ for $s$ such that $b(s)\neq 0$} \\
    \text{arbitrary} & \text{ for $s$ such that $b(s)= 0$}
    \end{cases}
\end{equation}
for each $m=1,2, ... , M$ and $s\in S$.
Note that $\tilde{\mathcal{P}}^*(m|s,b)$ is well-defined by our inclusion of the constraint that $\text{support}(\hat{b}^m) \subseteq \text{support}(\hat{b})$ in the definition of $\pi(\hat{b})$ in~\eqref{eq:def_pi_b}. Specifically, if $b(s)=0$, then by~\eqref{eq:p_star_alpha_expression}, it must hold that $\hat{b}^{z}(s)=0$ as well. Then, under the constraint in \eqref{eq:def_pi_b}, it follows that $\hat{b}^{m}(s)=\pi(\hat{b}^{z})(s)=0$.
Likewise, define $\tilde{\mathcal{A}}^*$ as
\begin{equation}
    \label{eq:def_a_tilde}
    \tilde{\mathcal{A}}^*(a|\hat{b}^m)=\begin{cases}
    \frac{1}{\alpha_b^m}\int_{N(\hat{b}^m)}\alpha_b^z \mathcal{A}^*(a|\hat{b}^z) dz  & \text{ if $\alpha_b^m \neq 0$} \\
    \text{arbitrary} & \text{ if $\alpha_b^m = 0$}
    \end{cases}
\end{equation}
for each $a\in A$ and $\hat{b}^m\in\hat{\mathcal{B}}$.

Notice that under the perception policy $\tilde{\mathcal{P}}^*$, a measurement $m$ is observed with probability
\[
\sum_s \tilde{\mathcal{P}}^*(m|s,b)b(s)=\sum_s \alpha_b^m \hat{b}^m(s)=\alpha_b^m.
\]
Moreover, when $m$ is observed, by Bayes' rule \eqref{eq:probUpdate}, the posterior belief becomes $\hat{b}^m$, since, by the construction of $\tilde{\mathcal{P}}^*$,
\begin{align*}
    \frac{\tilde{\mathcal{P}}^*(m|s,b)b(s)}{\sum_{s'}\tilde{\mathcal{P}}^*(m|s',b)b(s')}&=\frac{\alpha_b^m \hat{b}^m(s)/b(s)\cdot b(s)}{\alpha_b^m} \\
    &=\hat{b}^m(s).
\end{align*}
Therefore, $\tilde{\mathcal{P}}^*\in \mathcal{P}(b\rightarrow \hat{\mathcal{B}})$ as desired.

Having $(\mathcal{A}^*,\mathcal{P}^*)$ and $(\tilde{\mathcal{A}}^*,\tilde{\mathcal{P}}^*)$ defined, we now compare the right hand side of \eqref{eq:vi_gap_lhs} evaluated under $(\mathcal{A}^*,\mathcal{P}^*)$ and the right hand side of \eqref{eq:vi_gap_rhs} evaluated under $(\tilde{\mathcal{A}}^*,\tilde{\mathcal{P}}^*)$ term-by-term.

\subsection{Comparison of $R(b,\mathcal{P}^*)$ and $R(b,\tilde{\mathcal{P}}^*)$}

Using \eqref{eq:p_star_alpha_expression}, the perception cost $R(b,\mathcal{P}^*)$ is
\begin{align}
    R(b,\mathcal{P}^*)&=\sum_s \int_z \mathcal{P}^*(z|s,b)b(s)\log \frac{\mathcal{P}^*(z|s,b)}{\alpha_b^z}dz \nonumber \\
    &=\sum_s \int_z \alpha_b^z\hat{b}^z(s)\log \frac{\hat{b}^z(s)}{b(s)}dz \nonumber \\
    &=\int_z \alpha_b^z D(\hat{b}^z \| b)dz
    \label{eq:p_star_cost}
\end{align}
On the other hand, the perception cost $R(b,\tilde{\mathcal{P}}^*)$ is
\begin{subequations}
\label{eq:p_tilde_star_cost}
\begin{align}
    R(b,\tilde{\mathcal{P}}^*)&=\sum_s\sum_m \tilde{\mathcal{P}}^*(m|s,b)b(s) \log \frac{\tilde{\mathcal{P}}^*(m|s,b)}{\alpha_b^m} \nonumber \\
    &=\sum_s\sum_m \alpha_b^m \hat{b}^m(s) \log \frac{\hat{b}(s)^m}{b(s)} \label{eq:p_tilde_star_cost1} \\
    &=\sum_m \alpha_b^m D(\hat{b}^m \| b) \nonumber \\
    &=\sum_m \int_{N(\hat{b}^m)} \alpha_b^z D(\hat{b}^m \| b) dz \label{eq:p_tilde_star_cost2} \\
    &=\sum_m \int_{N(\hat{b}^m)} \alpha_b^z D(\pi(\hat{b}^z) \| b) dz \label{eq:p_tilde_star_cost3}\\
    &=\int_{Z} \alpha_b^z D(\pi(\hat{b}^z) \| b) dz  \label{eq:p_tilde_star_cost4}
\end{align}
\end{subequations}
Step \eqref{eq:p_tilde_star_cost1} follows by the construction of $\tilde{\mathcal{P}}^*$ in \eqref{eq:def_p_tilde}. The definition of $\alpha_{b}^{m}$ in \eqref{eq:def_alpha_m} is used to obtain \eqref{eq:p_tilde_star_cost2}. To obtain \eqref{eq:p_tilde_star_cost3}, recall that $\pi(\hat{b}^z)=\hat{b}^m$ for each $z\in N(\hat{b}^m)$. Finally, applying the fact that $\bigcup_{m=1}^M N(\hat{b}^m) = Z$ yields \eqref{eq:p_tilde_star_cost4}.

Comparing \eqref{eq:p_star_cost} and \eqref{eq:p_tilde_star_cost4}, we have then have that
\begin{align*}
    &\left|R(b,\mathcal{P}^*)-R(b,\tilde{\mathcal{P}}^*)\right| \\
    &=\left| \int_\mathcal{Z} \left(
    D(\hat{b}^z\|b)-D(\pi(\hat{b}^z)\|b)
    \right)dz\right|\\
    &\leq \int_\mathcal{Z} \left|
    D(\hat{b}^z\|b)-D(\pi(\hat{b}^z)\|b)
    \right| dz \\
    &\leq \max_{\hat{b}\in\Delta(S)} \left|
    D(\hat{b}\|b)-D(\pi(\hat{b})\|b)
    \right|
\end{align*}
The last expression can be upper bounded as follows:
\begin{subequations}
    \begin{align}
        & | D(\hat{b}\|b) - D(\pi(\hat{b})\|b) | \nonumber \\
        & \, = \Big|\sum_{s\in \overline{S}(b)}-\hat{b}(s) \log \frac{\hat{b}(s)}{b(s)} \nonumber \\
        & \qquad \qquad + \sum_{s\in \overline{S}(b)} \pi(\hat{b})(s) \log \frac{\pi(\hat{b})(s)}{b(s)} \Big| \label{eq:ubkl_1} \\
        & \, = \Big| \sum_{s\in \overline{S}(b)} -\hat{b}(s) \log \hat{b}(s) + \sum_{s\in \overline{S}(b)} \pi(\hat{b})(s) \log \pi(\hat{b})(s) \nonumber \\
        & \qquad \qquad + \sum_{s\in\overline{S}(b)} (\pi(\hat{b})(s) - \hat{b}(s)) \log b(s) \Big| \label{eq:ubkl_2} \\
        & \, = \Big| H(\hat{b}) - H(\pi(\hat{b})) \nonumber \\
        & \qquad \qquad + \sum_{s\in\overline{S}(b)} (\pi(\hat{b})(s) - \hat{b}(s)) \log b(s) \Big| \label{eq:ubkl_3} \\
        & \, \leq | H(\hat{b}) - H(\pi(\hat{b})) | \nonumber \\
        & \qquad \qquad + \Big| \sum_{s\in\overline{S}(b)} (\pi(\hat{b})(s) - \hat{b}(s)) \log b(s) \Big| \label{eq:ubkl_4} \\
        & \, \leq | H(\hat{b}) - H(\pi(\hat{b})) | + \hat{\epsilon} \Big| \sum_{s\in \overline{S}(b)} \log b(s) \Big| \label{eq:ubkl_5} \\
        & \, \leq \hat{\epsilon} |\log \hat{\epsilon} | |S|+ \hat{\epsilon} \Big| \sum_{s\in \overline{S}(b)} \log b(s) \Big|, \label{eq:ubkl_6}
    \end{align}
\end{subequations}
where \eqref{eq:ubkl_2} follows from standard properties of logarithms, \eqref{eq:ubkl_3} by the definition of the entropy function, \eqref{eq:ubkl_4} by the triangle inequality, \eqref{eq:ubkl_5} by the fact that $\|\hat{b} - \pi(\hat{b})\|_{\infty} \leq \hat{\epsilon}$, and \eqref{eq:ubkl_6} from the result of Lemma \ref{lem:upper_bound_entropy2}. Recall that the relative entropy is summed over only the support of $b$, denoted $\overline{S}(b)$. Thus, $\log b(s)$ is finite for all $b\in \mathcal{B}$ and $s \in \overline{S}(b)$.

\subsection{Comparison of $\mathbb{E}_b^{\mathcal{A}^*,\mathcal{P}^*}[C(\bs,\ba)]$ and $\mathbb{E}_b^{\tilde{\mathcal{A}}^*,\tilde{\mathcal{P}}^*}[C(\bs,\ba)]$}

Next, we compare the terms pertaining to $\mathbb{E}_b^{\mathcal{A},\mathcal{P}}[C(\bs,\ba)]$ in \eqref{eq:vi_gap_lhs} and \eqref{eq:vi_gap_rhs}.
Notice that
\begin{equation}
\label{eq:exp_c_ap_star}
\mathbb{E}_b^{\mathcal{A}^*,\mathcal{P}^*}[C(s,a)] 
  =\sum_{s,a} \left(\int_Z \alpha_b^z\mathcal{A}^*(a|\hat{b}^z)\hat{b}^z(s) dz\right) C(s,a).
\end{equation}
On the other hand,
\begin{subequations}
\label{eq:exp_c_ap_tilde}
\begin{align}
  &\mathbb{E}_b^{\tilde{\mathcal{A}}^*,\tilde{\mathcal{P}}^*}[C(\bs,\ba)] \nonumber \\
  &=\sum_{s,a} \sum_m \alpha_b^m \tilde{\mathcal{A}}^*(a|\hat{b}^m)\hat{b}^m(s) C(s,a) \nonumber \\
  &=\sum_{s,a} \sum_m \left(\int_{N(\hat{b}^m)} \alpha_b^z \mathcal{A}^*(a|\hat{b}^z)  dz \right)\hat{b}^m(s) C(s,a) \label{eq:exp_c_ap_tilde1}\\
  &=\sum_{s,a} \sum_m \left(\int_{N(\hat{b}^m)} \alpha_b^z \mathcal{A}^*(a|\hat{b}^z) \pi(\hat{b}^z)(s)  dz \right) C(s,a) \label{eq:exp_c_ap_tilde2}\\
  &=\sum_{s,a} \left(\int_Z \alpha_b^z \mathcal{A}^*(a|\hat{b}^z) \pi(\hat{b}^z)(s)  dz \right) C(s,a) \label{eq:exp_c_ap_tilde3}
\end{align}
\end{subequations}
Equality \eqref{eq:exp_c_ap_tilde1} follows from the definition of $\tilde{\mathcal{A}}^*(a|\hat{b}^m)$ in \eqref{eq:def_a_tilde}.
Equality \eqref{eq:exp_c_ap_tilde2} then holds by recalling that $\pi(\hat{b}^z)=\hat{b}^m$ for each $z\in N(\hat{b}^m)$. The fact that $\bigcup_{m=1}^M N(\hat{b}^m) = Z$ is used to obtain \eqref{eq:exp_c_ap_tilde3}.
Now, comparing \eqref{eq:exp_c_ap_star} and \eqref{eq:exp_c_ap_tilde3}, we have that
\begin{subequations}
\begin{align}
&\left| \mathbb{E}_b^{\mathcal{A}^*,\mathcal{P}^*}[C(\bs,\ba)]-\mathbb{E}_b^{\tilde{\mathcal{A}}^*,\tilde{\mathcal{P}}^*}[C(\bs,\ba)]\right| \nonumber \\
&=\left| \sum_{s,a} \left(\int_Z \alpha_b^z\mathcal{A}^*(a|\hat{b}^z)\hat{b}^z(s) dz\right) C(s,a) \right. \nonumber \\
&\qquad \left. -\sum_{s,a} \left(\int_Z \alpha_b^z \mathcal{A}^*(a|\hat{b}^z) \pi(\hat{b}^z)(s)  dz \right) C(s,a)\right| \nonumber \\
&\leq \sum_{s,a}\left|C(s,a)\right| \left(\int_Z \alpha_b^z\mathcal{A}^*(a|\hat{b}^z) |\hat{b}^z(s)-\pi(\hat{b}^z)(s)|dz \right)  \nonumber \\
&\leq \epsilon \sum_{s,a}\left|C(s,a)\right|\left(\int_Z \alpha_b^z\mathcal{A}^*(a|\hat{b}^z) dz \right)   \\
&\leq \epsilon \sum_{s,a}\left|C(s,a)\right|\left(\int_Z \alpha_b^z dz \right)   \\
&= \epsilon \sum_{s,a} \left|C(s,a)\right|.
\end{align}
\end{subequations}

\subsection{Comparison of $\mathbb{E}_b^{\mathcal{A}^*,\mathcal{P}^*} [V(b^{\bz,\ba})]$ and $\mathbb{E}_b^{\tilde{\mathcal{A}},\tilde{\mathcal{P}}^*} [V(b^{\bz,\ba})]$}

Finally, we compare the term $\mathbb{E}_b^{\mathcal{A},\mathcal{P}} [V(b^{\bz,\ba})]$ in \eqref{eq:vi_gap_lhs} and $\mathbb{E}_b^{\mathcal{A},\mathcal{P}} [V(b^{\bm,\ba})]$ in \eqref{eq:vi_gap_rhs}.
Notice that, under the policy $(\mathcal{P}^*, \mathcal{A}^*)$, the random variables $(\bz,\ba)$ are realized according to the probability distribution 
\[
\mathcal{A}^*(a|\hat{b}^z)\alpha_b^z=\mathcal{A}^*(a|\hat{b}^z)\sum_s \mathcal{P}^*(z|s,b)b(s).
\]
Similarly, under the policy $(\tilde{\mathcal{P}}^*, \tilde{\mathcal{A}}^*)$, the random variables $(\bm,\ba)$ are realized according to the probability distribution $\tilde{\mathcal{A}}^*(a|\hat{b}^m)\alpha_b^m$.
Therefore,
\begin{align}
    &\Big| \mathbb{E}_b^{\mathcal{A}^*,\mathcal{P}^*}[V_k(b^{\bz,\ba})]-\mathbb{E}_b^{\tilde{\mathcal{A}}^*,\tilde{\mathcal{P}}^*}[V_k(b^{\bm,\ba})]\Big| \nonumber \\
    &=\Big|\sum_a \int_Z \mathcal{A}^*(a|\hat{b}^z)\alpha_b^z V_k(b^{z,a})dz \nonumber \\
    &\qquad - \sum_a \sum_m \tilde{\mathcal{A}}^*(a|\hat{b}^m)\alpha_b^m V_k(b^{m,a})\Big| \nonumber \\
    &=\Big|\sum_a \int_Z \mathcal{A}^*(a|\hat{b}^z)\alpha_b^z V_k(b^{z,a})dz \nonumber \\
    &\qquad - \sum_a \sum_m \int_{N(\hat{b}^m)} \mathcal{A}^*(a|\hat{b}^z)\alpha_b^z V_k(b^{m,a})\Big|.  \label{eq:exp_v}
\end{align}
To obtain the second equality in \eqref{eq:exp_v}, we again use the definition of  $\tilde{\mathcal{A}}^*(a|\hat{b}^m)$ in \eqref{eq:def_a_tilde}.
Denote by $b^{m(z),a}$ the prior belief obtained when action $a$ is selected while in the posterior belief $\pi(\hat{b}^z)$. That is,
\[
b^{m(z),a}(s)=\sum_{s'} T(s|a,s')\pi(\hat{b}^z)(s').
\]
It is noteworthy that for each $s\in S$ and $z \in Z$,
\begin{align*}
    &\big| b^{z,a}(s)-b^{m(z),a}(s)\big|  \\
    &=\Big| \sum_{s'} T(s|a,s')\hat{b}^z(s') - \sum_{s'} T(s|a,s')\pi(\hat{b}^z)(s')\Big| \\
    &\leq \sum_{s'}\underbrace{T(s|a,s')}_{\leq 1} \underbrace{\big|\hat{b}^z(s)-\pi(\hat{b}^z)(s)\big|}_{\leq \hat{\epsilon}} \\
    &\leq \hat{\epsilon} |S|.
\end{align*}
Since $b^{m(z),a}=b^{m,a}$ for each $z\in N(\hat{b}^m)$, \eqref{eq:exp_v} can be upper bounded as follows:
\begin{align}
    \text{\eqref{eq:exp_v}}&=\Big|\sum_a \int_Z \mathcal{A}^*(a|\hat{b}^z)\alpha_b^z V_k(b^{z,a})dz \nonumber \\
    &\qquad - \sum_a  \int_Z\mathcal{A}^*(a|\hat{b}^z)\alpha_b^z V_k(b^{m(z),a})\Big| \nonumber \\
    &\leq \sum_a \int_Z \mathcal{A}^*(a|\hat{b}^z) \alpha_b^z\underbrace{\big| V_k(b^{z,a})-V_k(b^{m(z),a})\big|}_{\leq \hat{\delta}}dz \nonumber \\
    &\leq \hat{\delta} \sum_a \int_Z \mathcal{A}^*(a|\hat{b}^z) \alpha_b^z \label{eq:lipschitz} \\
    &\leq \hat{\delta}. \nonumber 
\end{align}
In step \eqref{eq:lipschitz}, we used the hypothesis \eqref{eq:continuity_hypothesis}.

\subsection{Summary} 
Summarizing (i), (ii) and (iii) above, we obtain 
$
|(TV)(b)-(\tilde{T}(V|_\mathcal{B}))(b)|\leq \epsilon
$
for each $b\in\mathcal{B}$. This completes the proof. $\Box$

\section{Additional information on 3-state example}
\label{appendix:three_state}

\begin{figure}
    \centering
    \begin{tikzpicture}[->, >=stealth', auto, semithick, node distance=2cm]
            \tikzstyle{every state}=[fill=white,draw=black,thick,text=black,scale=0.9]

\node[state] (s1) at (0,0) {\large{$s^{1}$}};
\node[state] (s2) at (3,0) {\large{$s^{2}$}};
\node[state] (s3) at (1.5,-2) {\large{$s^{3}$}};

\draw (s1) edge[out=20,in=160,->] (s2);
\draw (s2) edge[out=200,in=-20,->] (s1);
\draw (s1) edge[out=-40,in=110,->] (s3);
\draw (s3) edge[out=150,in=-80,->] (s1);
\draw (s2) edge[out=260,in=30,->] (s3);
\draw (s3) edge[out=70,in=220,->] (s2);
\draw (s1) edge[loop above] (s1);
\draw (s2) edge[loop above] (s2);
\draw (s3) edge[loop below] (s3);

\end{tikzpicture}
    \caption{3-state environment considered.}
    \label{fig:3StateExample}
\end{figure}
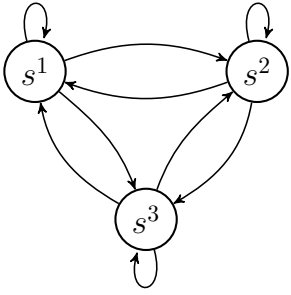

We consider the 3-state MDP shown in Fig.~\ref{fig:3StateExample}, where each state has three available actions. For notational convenience, introduce
\begin{equation*}
   \mathcal{T}^{a^{l}} = \begin{bmatrix}
        \mathcal{T}(s^{1}|s^{1},a^{l}) & \cdots & \mathcal{T}(s^{N}|s^{1},a^{l}) \\
        \vdots & \ddots & \vdots \\
        \mathcal{T}(s^{1}|s^{N},a^{l}) & \cdots & \mathcal{T}(s^{N}|s^{N},a^{l})
    \end{bmatrix}.
\end{equation*}
Using this notation, the 3-state example considered has the following transition dynamics:
\begin{align*}
    \mathcal{T}^{a^{1}} & = \begin{bmatrix} 
        0.1 & 0.9 & 0 \\
        0 & 0.1 & 0.9 \\
        0.5 & 0.5 & 0
        \end{bmatrix}, \, 
    \mathcal{T}^{a^{2}} = \begin{bmatrix}  
        0.1 & 0 & 0.9 \\
        0.9 & 0.1 & 0 \\
        0.5 & 0.5 & 0
    \end{bmatrix}, \\
    \mathcal{T}^{a^{3}} & = \begin{bmatrix}  
        0.998 & 0.001 & 0.001 \\
        0.001 & 0.998 & 0.001 \\
        0.001 & 0.001 & 0.998
    \end{bmatrix}. \nonumber
\end{align*}
\noindent
The objective of the agent is to avoid state $s^{3}$. To model this objective, we set the cost of taking any action while in $s^{3}$ to 1; i.e., $C(s^{3},a^{l})$$=$$1$ for all $l$$=$$1$$\ldots$$3$. We consider the problem of the agent over an infinite time horizon. To construct the set $\hat{\mathcal{B}}$, we uniformly grid the three-dimensional belief simplex using three different spacings. For the leftmost plot in Fig.~\ref{fig:increase_samples}, a spacing of $0.2$ was used, producing a total of $21$ posterior belief states and $63$ prior belief states. Similarly, for the center plot in Fig.~\ref{fig:increase_samples} a spacing of $0.1$ was used, yielding $62$ posterior belief states and $186$ prior belief states. Finally, the rightmost plot in Fig. 4 was produced using a spacing of $0.05$, which resulted in a total of $217$ posterior belief states and a corresponding set of $651$ prior belief states.

To obtain the values of prior belief states plotted we perform value iteration until convergence using values of $\gamma$$=$$5$ and $\beta$$=$$0.95$, wherein we solve (12) for each prior belief state and (13) for each posterior belief state at each iteration. To solve each linear program, we use the default linear program solver available in the \cite{MatlabOTB}.

\end{document}